\def\arxivmode{1}
\documentclass[10pt]{article}
\ifdefined\arxivmode\usepackage[preprint]{tmlr}\else\usepackage{tmlr}\fi

\usepackage[utf8]{inputenc}
\usepackage[T1]{fontenc}
\usepackage{hyperref}
\usepackage{url}
\usepackage{booktabs}
\usepackage{array}
\usepackage{amsfonts}
\usepackage{amsmath}
\usepackage{microtype}
\usepackage{graphicx}
\usepackage{acro}
\usepackage{float}
\graphicspath{{figures/}{generated/}}

\DeclareAcronym{MI}{short=MI, long={motor imagery}}
\DeclareAcronym{ERP}{short=ERP, long={event-related potential}}
\DeclareAcronym{MLP}{short=MLP, long={multilayer perceptron}}
\DeclareAcronym{FBCSP}{short=FBCSP, long={filter-bank common spatial pattern}}
\newcommand{\arm}[1]{\csname arm@#1\endcsname}
\expandafter\def\csname arm@pre\endcsname{pretrained}
\expandafter\def\csname arm@rand\endcsname{random-init}
\expandafter\def\csname arm@bp\endcsname{band power}

\newcommand{\numBciWBpMin}{0.514}
\newcommand{\numBciWBpMax}{0.516}
\newcommand{\numErpNFm}{6}

\newcommand{\numErpRandGePre}{1}
\newcommand{\numErpPreAboveRand}{5}

\newcommand{\numErpGapMaxPos}{+0.065}
\newcommand{\numErpGapMinPos}{+0.032}
\newcommand{\numSleepNFm}{6}

\newcommand{\numSleepCbramodGap}{+0.135}

\newcommand{\numSleepRandGePre}{3}

\newcommand{\numSleepAboveBp}{5}

\newcommand{\numEmoNFm}{6}

\newcommand{\numEmoRandGePre}{3}

\newcommand{\numEmoBelowBp}{6}

\newcommand{\numBciWNFm}{5}

\newcommand{\numBciWRandGePre}{2}

\newcommand{\numSleepAboveBpClear}{4}

\newcommand{\numCovSeeds}{3}

\newcommand{\numFtSeeds}{5}
\newcommand{\numFtCells}{30}
\newcommand{\numFtPos}{23}
\newcommand{\numFtPosTwoSd}{5}
\newcommand{\numFtNegTwoSd}{1}

\newcommand{\numFtPosTwoSdModels}{3}
\newcommand{\numFtPosTwoSdMaxPerModel}{2}

\newcommand{\numFtCbramodBciDelta}{+0.052}
\newcommand{\numFtCbramodBciDeltaSD}{0.030}

\newcommand{\numFtCbramodK}{5}

\newcommand{\numFtLabramTuslDelta}{+0.059}
\newcommand{\numFtLabramTuslDeltaSD}{0.075}

\newcommand{\numFtLabramK}{5}

\newcommand{\numFtEegptK}{3}
\newcommand{\numFtLunaApavaDelta}{$-$0.290}
\newcommand{\numFtLunaApavaDeltaSD}{0.035}

\newcommand{\numFtLunaK}{3}

\newcommand{\numFtLumambaK}{3}

\newcommand{\numFtOursK}{4}

\newcommand{\numFtCbramodBciSeedOne}{+0.102}
\newcommand{\numMiCells}{12}
\newcommand{\numMiAboveBp}{0}
\newcommand{\numOrigOursLeePeak}{0.708}
\newcommand{\numOrigOursLeeFinal}{0.507}
\newcommand{\numOrigOursLeeBp}{0.741}

\newcommand{\numOrigBciBpMin}{0.484}
\newcommand{\numOrigBciBpMax}{0.491}

\newcommand{\numSelCells}{36}
\newcommand{\numSelPenaltyMean}{+0.034}
\newcommand{\numSelPenaltyMax}{+0.191}
\newcommand{\numSelFlipThr}{0.01}
\newcommand{\numSelFlip}{9}
\newcommand{\numSelHonest}{11}
\newcommand{\numSelLumambaApavaGapVal}{$-$0.139}
\newcommand{\numSelLumambaApavaGapTest}{+0.006}
\newcommand{\numSelOursLeeGapVal}{+0.202}

\newcommand{\numSelLunaApavaValGap}{+0.104}

\newcommand{\numBudgetGrowThr}{0.02}
\newcommand{\numAttnCells}{6}

\newcommand{\numAttnDMin}{$-$0.069}
\newcommand{\numAttnDMax}{$-$0.007}
\newcommand{\numTfiveDeltaAsym}{$-$0.272}
\newcommand{\numTfiveDeltaLrLow}{$-$0.184}
\newcommand{\numTfiveDeltaLrMid}{$-$0.287}
\newcommand{\numTfiveDeltaLrHigh}{$-$0.124}
\newcommand{\numTfiveSeeds}{5}
\newcommand{\numMlpCells}{18}
\newcommand{\numMlpRecoverMax}{1}
\newcommand{\numMlpSeeds}{3}
\newcommand{\numBudgetCells}{18}

\newcommand{\numBudgetLunaApavaFullGap}{+0.011}
\newcommand{\numBudgetGrowMean}{5}
\newcommand{\numBudgetGrowMeanPos}{2}
\newcommand{\numBudgetLowFrac}{10}
\newcommand{\numBciWithinMinusCrossMin}{$-$0.003}
\newcommand{\numBciWithinMinusCrossMax}{+0.044}
\newcommand{\numBciWithinMaxPeak}{0.370}
\newcommand{\numBciCrossMaxPeak}{0.338}
\newcommand{\numBciChance}{0.25}
\newcommand{\numEmoChance}{0.125}
\newcommand{\numFtLunaApavaProbeDelta}{+0.040}
\newcommand{\numTfourCells}{15}
\newcommand{\numTfourPenMean}{+0.031}
\newcommand{\numTfourPenMin}{$-$0.019}
\newcommand{\numTfourPenMax}{+0.148}
\newcommand{\numTfourNeg}{2}
\newcommand{\numTfourCanonKept}{4}

\newcommand{\numModeFrozenOnly}{3}

\newcommand{\numModeThr}{0.01}
\newcommand{\numModeFrozenOnlyStable}{3}
\newcommand{\numModeBothPosLo}{12}
\newcommand{\numModeBothPosHi}{17}

\newcommand{\numEegnetLee}{0.796}
\newcommand{\numEegnetLeeSD}{0.046}
\newcommand{\numEegnetTuab}{0.863}
\newcommand{\numEegnetTuabSD}{0.009}

\newcommand{\numRandDrawCells}{18}
\newcommand{\numRandDrawDiffMean}{0.030}
\newcommand{\numRandDrawDiffMax}{0.115}
\newcommand{\numRandDrawMaxCell}{LaBraM on APAVA}
\newcommand{\numRandDrawMaxA}{0.857}
\newcommand{\numRandDrawMaxB}{0.741}
\newcommand{\numFtLabramTuslSeedOne}{$-$0.060}
\newcommand{\numDevRawProbe}{0.599}
\newcommand{\numDevVfiveEncProbe}{0.509}
\newcommand{\numDevQStdStart}{0.020}
\newcommand{\numDevQStdEnd}{0.023}
\newcommand{\numDevQSteps}{45K}
\newcommand{\numDevVfiveStop}{55K}
\newcommand{\numDevVeightEncProbe}{0.595}
\newcommand{\numDevConvProbe}{0.538}
\newcommand{\numDevSincProbe}{0.715}
\newcommand{\numDevSincBp}{0.691}
\newcommand{\numDevTrajApavaStep}{25K}
\newcommand{\numDevTrajApavaPeak}{+0.127}
\newcommand{\numDevTrajApavaEnd}{+0.036}
\newcommand{\numDevTrajBciStep}{5K}
\newcommand{\numDevTrajBciPeak}{+0.014}
\newcommand{\numDevTrajBciEnd}{$-$0.034}

\newcommand{\numDevTrajTuabEnd}{+0.000}
\newcommand{\numDevTrajLeeStep}{15K}
\newcommand{\numDevTrajLeePeak}{+0.024}
\newcommand{\numDevTrajLeeEnd}{$-$0.034}
\newcommand{\numMdlFilters}{64}
\newcommand{\numMdlSincKernel}{501}
\newcommand{\numMdlPatch}{25}
\newcommand{\numMdlPatchMs}{125}
\newcommand{\numMdlPatches}{40}

\newcommand{\numMdlSrate}{200}

\newcommand{\numMdlWidth}{256}
\newcommand{\numMdlDepth}{8}

\newcommand{\numMdlPredDepth}{6}
\newcommand{\numMdlPredWidth}{768}

\newcommand{\numMdlMaskRatio}{0.60}
\newcommand{\numMdlMaskMinSec}{0.50}
\newcommand{\numMdlMaskMaxSec}{1.00}
\newcommand{\numMdlChanBlockMin}{1}
\newcommand{\numMdlChanBlockMax}{4}
\newcommand{\numMdlSigProj}{1024}

\newcommand{\numMdlWindowSec}{5}

\newcommand{\numMdlSigT}{3}

\newcommand{\numRunMaxSteps}{100{,}000}
\newcommand{\numRunWarmupFrac}{0.1}
\newcommand{\numRunLr}{$5{\times}10^{-5}$}

\newcommand{\numRunLambda}{1.0}

\newcommand{\numRunEffBatch}{512}
\newcommand{\numFeFilters}{64}
\newcommand{\numFeBelowOne}{64}
\newcommand{\numFeDesignLo}{2}
\newcommand{\numFeDesignHi}{100}
\newcommand{\numFeEdgeMin}{0.24}
\newcommand{\numFeEdgeMax}{0.73}

\newcommand{\numFeShareMuBetaIn}{14}

\newcommand{\numFeShareMuBetaCoded}{0.58}

\newcommand{\numRankSlowFirstOf}{all four}
\newcommand{\numRankSpecLastOf}{all three}
\newcommand{\numRankOursLee}{0.709}
\newcommand{\numRankNextLee}{0.570}
\newcommand{\numRankOursErp}{0.795}
\newcommand{\numRankNextErp}{0.768}
\newcommand{\numFtPreCbramodLee}{0.848}
\newcommand{\numFtPreOursLee}{0.515}
\newcommand{\numFtPairCells}{30}
\newcommand{\numFtPairCiPos}{7}
\newcommand{\numFtPairCiNeg}{2}
\newcommand{\numFtPairBh}{5}
\newcommand{\numFtPairBhTuh}{2}

\newcommand{\numBudOrigCells}{18}
\newcommand{\numBudOrigFlips}{3}
\newcommand{\numBudOrigFlipMax}{0.085}
\newcommand{\numNatRateMatch}{4}
\newcommand{\numNatModels}{5}
\newcommand{\numBciWEegnet}{0.552}
\newcommand{\numBciWEegnetSD}{0.008}
\newcommand{\numBciWEegnetSeeds}{5}
\newcommand{\numBciWFbcsp}{0.421}
\newcommand{\numEmoNullK}{500}
\newcommand{\numEmoNullSeeds}{3}
\newcommand{\numEmoNullMeanMin}{0.117}
\newcommand{\numEmoNullMeanMax}{0.118}
\newcommand{\numEmoNullQMax}{0.136}
\newcommand{\numEmoNullBpMin}{0.168}
\newcommand{\numEmoNullBpMax}{0.220}
\newcommand{\numEmoNullP}{0.002}
\newcommand{\eqFrontend}{%
\begin{equation}
  g_k[n]=\frac{w[n]}{2(\tilde f^{\mathrm{hi}}_k-\tilde f^{\mathrm{lo}}_k)}\Big(2\tilde f^{\mathrm{hi}}_k\,\mathrm{sinc}\big(2\pi\tilde f^{\mathrm{hi}}_k\,\tfrac{n}{f_s}\big)-2\tilde f^{\mathrm{lo}}_k\,\mathrm{sinc}\big(2\pi\tilde f^{\mathrm{lo}}_k\,\tfrac{n}{f_s}\big)\Big),\qquad \tilde f_k=f_k/f_s .
  \label{eq:frontend}
\end{equation}}
\newcommand{\eqTokens}{%
\begin{equation}
  \mathbf x_{c,t}=W\,\mathbf u_{c,t}+\mathrm{MLP}\big([\sin(\mathbf r_c^{\top}\mathbf B);\,\cos(\mathbf r_c^{\top}\mathbf B)]\big)+\mathbf e_t,
  \qquad \mathbf u_{c,:}=\mathrm{Conv}\big([\,g_1*s_c,\dots,g_K*s_c\,]\big).
  \label{eq:tokens}
\end{equation}}
\newcommand{\eqPred}{%
\begin{equation}
  \tilde{\mathbf z}_i=\begin{cases}\mathbf m, & i\in\mathcal M\\ \mathbf z_i, & i\notin\mathcal M\end{cases},\qquad
  \mathcal{L}_{\mathrm{pred}}=\frac{1}{|\mathcal M|}\sum_{i\in\mathcal M}\Big\lVert h'\big(p_\phi(\tilde{\mathbf z}+\mathbf e^{\mathrm{pos}})\big)_i-\mathrm{sg}\big(h(\mathbf z)_i\big)\Big\rVert_2^2,
  \qquad \mathbf z=f_\theta(\mathbf x).
  \label{eq:pred}
\end{equation}}
\newcommand{\eqObjective}{%
\begin{equation}
  \begin{aligned}
  \mathcal{L}&=\mathcal{L}_{\mathrm{pred}}+\lambda\,\mathrm{SIGReg}\big(h(\mathbf z)\big),\\
  \mathrm{SIGReg}(Y)&=\frac{1}{S\,A}\sum_{s=1}^{S}\sum_{a=1}^{A}N\!\int_{-T}^{T}\! \exp(-t^2/2)\,\Big|\frac{1}{N}\sum_{b=1}^{N}\exp(\mathrm{i}\,t\,\mathbf v_a^{\top}\mathbf y_{s,b})-\exp(-t^2/2)\Big|^2 dt .
  \end{aligned}
  \label{eq:objective}
\end{equation}}
\newcommand{\eqProbe}{%
\begin{equation}
  \hat s=\arg\max_{s}\,M^{\mathrm{val}}_{s},\qquad R=M^{\mathrm{test}}_{\hat s},\qquad
  G=M^{\mathrm{test}}_{\hat s}-M^{\mathrm{test}}_{s_{\mathrm{out}}},\qquad
  \pi=\max_{s}M^{\mathrm{test}}_{s}-M^{\mathrm{test}}_{\hat s}.
  \label{eq:probe}
\end{equation}}

\title{Controlled Evaluation of EEG Foundation Models:\\Random Initialisation, Band-Power and Validation-Selected Probes}
\author{\name Mingyu Jeon \email jkmcoma7@gmail.com \\
      \addr Modulabs, Seoul, Republic of Korea
      \AND
      \name Suwan Cho \email cho.suwan96@gmail.com \\
      \addr Modulabs, Seoul, Republic of Korea
      \AND
      \name Jae Young Suh \email tjwodud04@gmail.com \\
      \addr Modulabs, Seoul, Republic of Korea}

\ifdefined\arxivmode\newcommand{\releaseurl}{ at \url{https://github.com/fbdeme/eeg-fm-controlled-audit}}\else\newcommand{\releaseurl}{}\fi

\def\month{10}
\def\year{2026}
\def\openreview{\url{https://openreview.net/forum?id=XXXX}}

\begin{document}
\maketitle


\begin{abstract}
EEG foundation models report downstream gains, yet their protocols seldom include three controls.
These are the same encoder untrained, spectral features fitted on the same data, and a layer picked without looking at test scores.
We therefore evaluate five publicly released EEG foundation models and an in-house JEPA-style model under one protocol with all three controls and several seeds, on five benchmarks and four added task families.
The controls change the reading in three ways.
First, picking the probed layer by its test score raises the layer-wise peak by an average of \numSelPenaltyMean\ and reverses the peak-versus-output gap in \numSelFlip\ of \numSelCells\ cells.
Second, with the layer picked on validation, random initialisation matches or beats pretraining for half of the models on sleep and emotion, and no model beats a five-band log-power baseline on emotion or motor imagery.
Third, after full fine-tuning only \numFtPosTwoSd\ of \numFtCells\ pairs gain beyond twice their seed spread, and the one clear loss of \numFtLunaApavaDelta\ AUROC persists under matched learning rates.
From these results we draw four reporting practices, and we release every cell for re-analysis.

\end{abstract}
\acresetall  


\section{Introduction}
\label{sec:intro}

Recent EEG foundation models such as LaBraM \citep{jiang2024labram}, EEGPT \citep{wang2024eegpt}, CBraMod \citep{wang2025cbramod}, LUNA \citep{doner2025luna} and LuMamba \citep{broustail2026lumamba} each report downstream gains under their own protocol, with its own splits, heads, schedules and hyperparameters.
These protocols differ, so the published numbers cannot be compared across papers, and none of the papers evaluates the other models under its own protocol \citep{jiang2024labram,wang2025cbramod,doner2025luna}.
Three controls are missing from most of these reports.
The three are a matched encoder of the same architecture at random initialisation, a band-power baseline on the same split, and a probed layer chosen on validation instead of on test (Section~\ref{sec:protocols}).
Each is standard in representation-learning evaluation \citep{alain2017probes,musgrave2020metric,pineau2018reproducibility}.

Such controls have changed conclusions before.
In other fields they shrank the reported gains, as metric-learning improvements nearly vanished under one fixed protocol \citep{musgrave2020metric} and better ImageNet models did not always transfer better \citep{kornblith2019better}.
The same happened in time-series forecasting, where linear models rivalled transformer forecasters \citep{zeng2023transformers} and the pretrained language model could be removed from language-model forecasters without loss \citep{tan2024llmts}.
Whether EEG foundation models keep their gains once all three controls are in place is the question this paper measures.

We came to this question from the model side, after building an in-house JEPA-style model for arbitrary montages \citep{assran2023ijepa,balestriero2025lejepa,maes2026lewm} and pretraining it on the REVE corpus \citep{elouahidi2025reve} (Figure~\ref{fig:arch}).
Its frozen output-layer probe on Lee2019 \ac{MI} scored \numOrigOursLeeFinal\ AUROC, below the \numOrigOursLeeBp\ of a five-band log-power baseline, \arm{bp}, computed on the same split \citep{lee2019openbmi}.
Probing every internal stage then located the loss, because an early front-end stage reached \numOrigOursLeePeak\ and the later stages lost that signal again (Figure~\ref{fig:layerwise}, left).
That observation turned the project into a controlled evaluation of five public EEG foundation models.
The in-house model stays in the panel as one row, under the same adapter interface, splits, probe and hyperparameters as the others.

The paper makes three contributions.
First, it evaluates six models under all three controls, covering \numFtCells\ model-dataset pairs under frozen probing and full fine-tuning at \numFtSeeds\ seeds (Section~\ref{sec:setup}).
It extends them with four task families (\ac{ERP}, sleep, emotion and within-subject \ac{MI}) at \numCovSeeds\ seeds wherever the seed changes the result, with LuMamba absent from within-subject \ac{MI}.
Second, it reports what survives the controls: test selection inflates layer-wise peaks and the band-power baseline is rarely beaten (Section~\ref{sec:results}).
It also finds that random initialisation often matches pretraining under a frozen probe and that most fine-tuning gains sit within seed noise (Sections~\ref{sec:random} and~\ref{sec:fullft}).
Third, it describes the diagnostic protocol that exposed the in-house failure and turns it into reporting practices for evaluating EEG foundation models (Sections~\ref{sec:diagnostic} and~\ref{sec:discussion}).
One distinction runs through all three.
Representation quality, read by a frozen probe, is kept apart from trainability, the fine-tuning gain over random initialisation (Section~\ref{sec:protocols}).
Performance after per-model tuning is a third notion, which a shared protocol bounds from below and does not measure (Section~\ref{sec:fullft}).


\section{Related work}
\label{sec:related}

EEG foundation models differ in objective and input handling: LaBraM predicts masked codes from a learned neural tokenizer \citep{jiang2024labram}, and CBraMod reconstructs masked patches with criss-cross attention \citep{wang2025cbramod}.
EEGPT likewise combines masked prediction with reconstruction \citep{wang2024eegpt}, LUNA compresses channels through learned queries \citep{doner2025luna}, and LuMamba pairs a joint-embedding objective with a state-space backbone \citep{broustail2026lumamba}.
Earlier and parallel lines include contrastive pretraining \citep{kostas2021bendr}, multi-modal biosignal transformers \citep{yang2023biot}, generative pretraining \citep{cui2024neurogpt} and signal-level joint-embedding prediction \citep{guetschel2024sjepa}.
These papers compare against published numbers from other protocols, whereas we run every model ourselves under one protocol with the three controls (Appendix~\ref{app:models}).

Controlled re-evaluations are an established genre: reproducibility checklists \citep{pineau2018reproducibility}, re-labelled benchmarks \citep{beyer2020imagenet}, and fixed-protocol comparisons \citep{musgrave2020metric,kornblith2019better}.
Benchmarks of EEG foundation models already report that large models gain little over compact deep architectures \citep{lee2025brainwave}, standardise comparisons across paradigms \citep{xiong2026eegfmbench}, and find specialist models trained from scratch competitive \citep{liu2026eegfmcompass}.
A second tool comes from probing, where linear probes read how much of a label is linearly decodable at each layer \citep{alain2017probes} and masked autoencoders rank differently under linear probing and fine-tuning \citep{he2022mae}.
We apply that genre to EEG and add the layer-selection control, since a layer-wise probe reports as many candidate values as there are layers \citep{alain2017probes}.

The baselines a foundation model has to beat are older than the models.
Classical EEG decoding rests on band power in the canonical frequency bands \citep{babiloni2009ad,pfurtscheller1999erd}, on compact convolutional networks such as EEGNet \citep{lawhern2018eegnet}, and for \ac{MI} on filter-bank spatial filters \citep{ang2008fbcsp}.
We use all three on the same splits as the foundation models, so that a gain is measured against the simplest method that could have produced it (Table~\ref{tab:classical}).


\section{Methodology}
\label{sec:method}

\subsection{Models}
\label{sec:models}

This section fixes what is compared, on which data, and under which protocols.
The panel pairs the in-house model with five public foundation models, LaBraM \citep{jiang2024labram}, CBraMod \citep{wang2025cbramod}, EEGPT \citep{wang2024eegpt}, LUNA \citep{doner2025luna} and LuMamba \citep{broustail2026lumamba}, all loaded from checkpoints pinned to a repository revision (Table~\ref{tab:models}).
The six models also differ in how raw channels enter the encoder (Appendix~\ref{app:typology}).
That contrast is only a hypothesis here, because in a panel of six it is confounded with corpus and implementation maturity (Appendix~\ref{app:typology}).

\subsection{The in-house model}
\label{sec:inhouse}

\begin{figure}[t]\centering
\includegraphics[width=\textwidth]{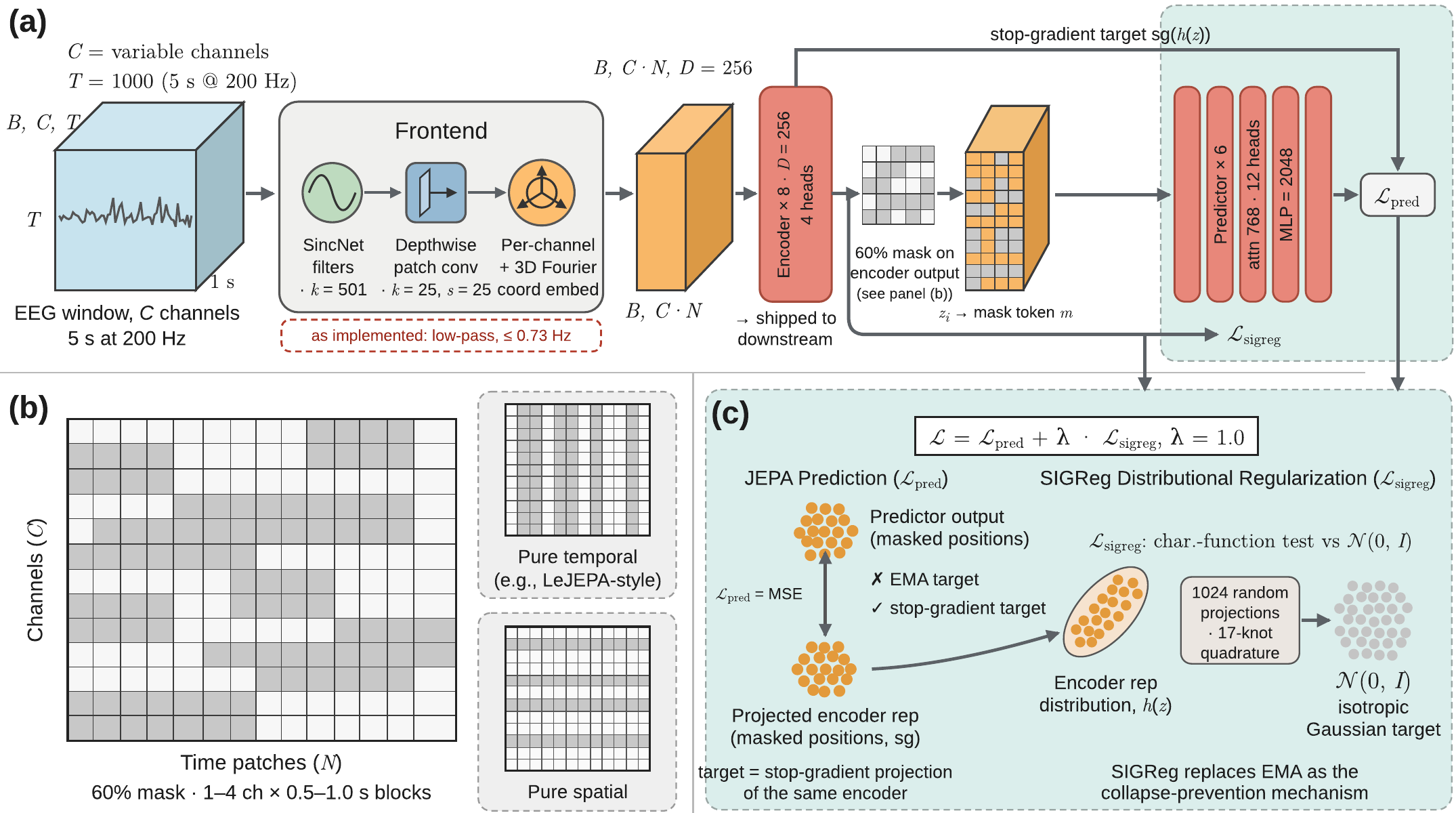}
\caption{The in-house model as released. Each channel passes through the sinc front-end (Eq.~\ref{eq:frontend}), which acts as a low-pass filter below one hertz as implemented, and a patch convolution; tokens add electrode-position and temporal embeddings (Eq.~\ref{eq:tokens}). The encoder reads the unmasked window, masking replaces its outputs by a mask token before the predictor, and the projected prediction is regressed onto the stop-gradient projection of the encoder output, with SIGReg on that projection (Eqs.~\ref{eq:pred} and~\ref{eq:objective}). Only the encoder is evaluated downstream.}
\label{fig:arch}
\end{figure}

Because its development is part of the results, the model this paper evaluates is described here in full.
The in-house model is a joint-embedding predictive architecture that replaces the averaged target encoder by the SIGReg regulariser \citep{assran2023ijepa,balestriero2025lejepa,maes2026lewm}, and Figure~\ref{fig:arch} shows it as released.
Its evaluated checkpoint is the one saved after 25{,}000 optimiser steps on REVE, chosen by the development trajectory of Section~\ref{sec:casea} \citep{elouahidi2025reve}.
That run used $\lambda=\numRunLambda$ in Eq.~\ref{eq:objective}, AdamW with a peak learning rate of \numRunLr, warm-up over a fraction \numRunWarmupFrac\ of a \numRunMaxSteps-step cosine schedule, and \numRunEffBatch\ windows per step.

Each channel $s_c$ first passes through $K=\numMdlFilters$ sinc filters of \numMdlSincKernel\ taps, each meant as a band-pass filter with a learnable lower cut-off and bandwidth as in SincNet \citep{ravanelli2019sincnet}.
Filter $g_k$ multiplies a Hamming window $w[n]$ over taps $n$ with the difference of two sinc functions at cut-offs $f^{\mathrm{lo}}_k<f^{\mathrm{hi}}_k$, normalised to $\tilde f^{\mathrm{lo}}_k$ and $\tilde f^{\mathrm{hi}}_k$ by the sampling rate $f_s$: \eqFrontend
This implementation divides both the cut-offs $f_k$ and the time index $n$ by the sampling rate $f_s$, where SincNet divides only one of them, so a filter designed for $f_k$\,Hz passes $f_k/f_s$\,Hz.
On the released checkpoint \numFeBelowOne\ of the \numFeFilters\ filters, designed to cover \numFeDesignLo\ to \numFeDesignHi\,Hz, act as low-pass filters with upper $-3$\,dB edges between \numFeEdgeMin\ and \numFeEdgeMax\,Hz (Appendix~\ref{app:frontend}).
The front-end output of Lee2019 windows therefore keeps \numFeShareMuBetaCoded\% of its spectral power in the 8--30\,Hz mu and beta bands, against \numFeShareMuBetaIn\% at its input.
The model thus sees little of the rhythms that motor imagery modulates \citep{pfurtscheller1999erd}.

A convolution with kernel and stride \numMdlPatch\ then turns each filtered channel $c$ into features $\mathbf u_{c,t}$ for \numMdlPatches\ patches $t$ of \numMdlPatchMs\,ms.
The token $\mathbf x_{c,t}$ adds a learned projection $W$ of these features, a \ac{MLP} embedding of random Fourier features $\mathbf B$ of electrode position $\mathbf r_c$ \citep{tancik2020fourier}, and a temporal embedding $\mathbf e_t$: \eqTokens
These embeddings are the only positional information, since the transformer blocks add none, and a cross-channel attention layer present in the code is switched off in the released configuration.

The \numMdlDepth-layer transformer encoder $f_\theta$ of width \numMdlWidth\ reads all tokens $\mathbf x$ of the unmasked window and returns $\mathbf z=f_\theta(\mathbf x)$.
That output is masked only afterwards: the outputs $\mathbf z_i$ at the masked positions $i\in\mathcal M$ are replaced by a learned token $\mathbf m$.
This set $\mathcal M$ covers a fraction \numMdlMaskRatio\ of the tokens in blocks of \numMdlMaskMinSec\ to \numMdlMaskMaxSec\,s and \numMdlChanBlockMin\ to \numMdlChanBlockMax\ channels.
On that masked output, a \numMdlPredDepth-layer predictor $p_\phi$ with attention width \numMdlPredWidth\ then fills in the masked positions of $\tilde{\mathbf z}$, to which learned position embeddings $\mathbf e^{\mathrm{pos}}$ are added.
Its output passes through a projection $h'$ and is regressed onto the projection $h(\mathbf z)$ under a stop-gradient $\mathrm{sg}$, with no averaged target encoder, and $\mathcal{L}_{\mathrm{pred}}$ averages over a batch's masked tokens: \eqPred

SIGReg acts on $Y=h(\mathbf z)$, whose entry $\mathbf y_{s,b}$ is token position $s$, one of $S$, of window $b$ in a batch of $N$ windows.
It compares, per position, the empirical characteristic function along $A=\numMdlSigProj$ random unit directions $\mathbf v_a$, redrawn every step, with a standard Gaussian's on $t\in[-T,T]$, $T=\numMdlSigT$ \citep{balestriero2025lejepa}.
In sum, the total loss $\mathcal{L}$ adds this term to $\mathcal{L}_{\mathrm{pred}}$ with weight $\lambda$: \eqObjective
Because the encoder has already attended to the masked positions, the outputs the predictor sees carry information about its targets (Eq.~\ref{eq:pred}).
That may make the prediction task easier than masking the input would \citep{assran2023ijepa}.

\subsection{Benchmarks and seed semantics}
\label{sec:setup}

\begin{table}[t]\centering\small
\caption{Original benchmarks: task, classes, windows per split (five seconds at 200\,Hz, read from the evaluation outputs) and the split rule. TUSL and TUAB draw a new subject-disjoint split for every seed (sizes shown for the first seed); the other three keep one split. The TUSL positive class is rare, so AUROC is reported.}
\label{tab:datasets}
\begin{tabular}{l>{\raggedright\arraybackslash}p{3.3cm}ccc>{\raggedright\arraybackslash}p{3.6cm}}
\toprule
Dataset & Task & classes & train / val / test & metric & split \\
\midrule
APAVA & Alzheimer's disease vs control & 2 & 347 / 157 / 159 & AUROC & fixed, four test subjects \\
BCI IV 2a & motor imagery & 4 & 2,880 / 1,152 / 1,152 & accuracy & fixed, subjects 1--5 / 6--7 / 8--9 \\
Lee2019 & motor imagery & 2 & 2,080 / 520 / 2,600 & AUROC & fixed, within-subject halves \\
TUSL & slowing vs other & 2 & 12,187 / 1,050 / 1,503 & AUROC & subject-disjoint, per seed \\
TUAB & abnormal vs normal & 2 & 12,866 / 3,192 / 3,942 & AUROC & subject-disjoint, per seed \\
\bottomrule
\end{tabular}
\end{table}

The original matrix covers Alzheimer's disease on APAVA \citep{escudero2006multiscale}, motor imagery on BCI IV 2a \citep{tangermann2012bci} and Lee2019 \citep{lee2019openbmi}, TUH slowing events (TUSL) and abnormal TUAB recordings \citep{obeid2016tuh,lopez2017tuab}.
All five are cut into \numMdlWindowSec-second windows at \numMdlSrate\,Hz under one preprocessing chain (Table~\ref{tab:datasets}).
These five are extended by four task families on data outside TUH, namely P300 detection on BNCI2014-009, five-class sleep staging on ISRUC, eight-class emotion recognition on FACED, and within-subject \ac{MI} on BCI IV 2a.
Here the public FACED release we read carries eight of the nine FACED emotion categories, without the neutral one, so the emotion task is eight-class \citep{chen2023faced}.
Each added task is reduced to one five-second window, the centre of each labelled epoch for sleep and emotion and a zero-padded window for the sub-second P300 epochs.
The reduction is the same for every model and for \arm{bp}.
Then each adapter passes the windows to its checkpoint in the input format that checkpoint's code expects, which matches pretraining only in part (Table~\ref{tab:native}).
Of the \numNatModels\ public models, \numNatRateMatch\ receive their pretraining sampling rate, while EEGPT receives the rate of the released port we load rather than that of its original release.
These differences extend to the window lengths of CBraMod and LaBraM and, for several models, to channel layout, reference or scaling; LUNA, for example, used bipolar, z-scored input for its TUEG pretraining.

A seed does not vary the same thing on every benchmark.
On sleep, emotion, TUSL and TUAB each seed draws a new subject-disjoint split, so every reported spread includes the variation between splits (Table~\ref{tab:datasets}).
The other benchmarks, APAVA, BCI IV 2a, Lee2019 and the two within-subject protocols (P300 and \ac{MI}), keep one split, so there only the random initialisation and the fine-tuning run change with the seed (Table~\ref{tab:datasets}).
On those fixed splits the frozen probe of a pretrained checkpoint is deterministic and is reported as a point estimate rather than with a zero-width error bar (Appendix~\ref{app:coverage}).

\subsection{Controlled protocols}
\label{sec:protocols}

\begin{figure}[t]\centering
\includegraphics[width=\textwidth]{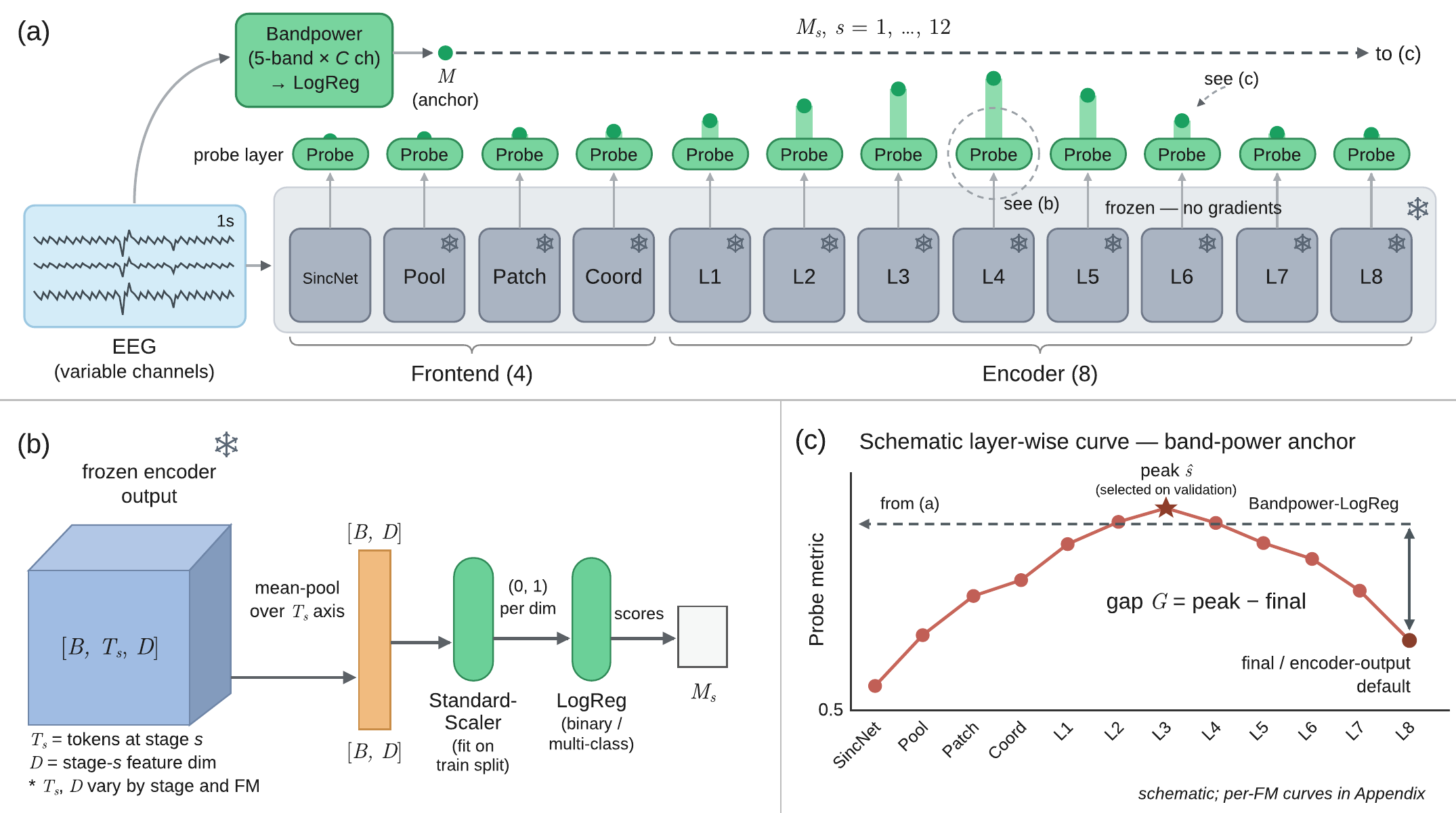}
\caption{Layer-wise probe protocol. (a) A linear probe at every internal stage of the frozen encoder (twelve for the in-house model), with the \arm{bp} anchor on the same split. (b) Per-stage mechanics: mean-pool over tokens, standardise on the training split, logistic regression. (c) The gap $G$ is the peak minus the output stage, with the peak $\hat s$ selected on validation (Eq.~\ref{eq:probe}; measured curves in Figure~\ref{fig:layerwise}).}
\label{fig:probe}
\end{figure}

Each model is wrapped in an adapter that exposes its internal stages; at each stage we mean-pool the tokens, standardise on the training split and fit a logistic-regression probe (Figure~\ref{fig:probe}).
Of the probed stages, the reported stage is the one with the best validation metric, scored once on test, and the test-argmax value is kept only to measure how much test selection would have added.
For a cell with probed stages $s$, output stage $s_{\mathrm{out}}$ and probe metric $M$, the reported value $R$ at the validation-selected stage $\hat s$, the peak-minus-output gap $G$ and the selection penalty $\pi$ are: \eqProbe
Two references accompany every probe.
Every \arm{pre} cell, the pretrained checkpoint, has a \arm{rand} twin, the same architecture at random initialisation with the library's default initialiser, probed identically.
The second reference, \arm{bp}, is the five-band log-power baseline: log band power per channel, followed by the same logistic regression on the same split.
It is computed on each model's own preprocessed input, so every model meets its matched baseline (Appendix~\ref{app:constants}).

Full fine-tuning trains each model with its own head under one canonical schedule.
The gain $\Delta$ is the \arm{pre} minus \arm{rand} test metric at the best-validation epoch, averaged over \numFtSeeds\ seeds with the population standard deviation.
Within that schedule, the learning rate is lower for \arm{pre} than for \arm{rand}, because a shared rate under-trains the random twin and biases $\Delta$ upward; Section~\ref{sec:fullft} re-runs the most negative cell under matched rates.
This single schedule gives a fair-comparison lower bound on each model's best, and Section~\ref{sec:fullft} measures the gap with a per-model learning-rate grid \citep{musgrave2020metric}.
To read $\Delta$ against its seed noise, we call a mean gain clear when it exceeds twice its seed standard deviation, a descriptive bar across \numFtCells\ cells.
It is a heuristic and not a corrected test, because with \numFtSeeds\ seeds a rank test cannot reach conventional significance in any single cell \citep{pineau2018reproducibility}.

\subsection{Diagnostic protocol}
\label{sec:diagnostic}

Everything above was assembled while diagnosing the in-house model, where every architectural iteration was kept or dropped by the same stage-wise probes (Section~\ref{sec:casea}).
That diagnosis also produced two further tools.
First, each model-dataset pair is labelled by the signs of its frozen-probe gain and its fine-tuning gain, which separates a representation defect (both negative) from a trainability defect (frozen positive, fine-tuning negative).
Second, the same probes run on every checkpoint saved during pretraining, so that a run can be stopped when its downstream probes turn down rather than when its training loss stops falling.
By contrast, on the five public models the protocol is applied after the fact as an evaluator; we did not use it during their development and make no claim that it would have changed it.


\section{Experiments}
\label{sec:results}

The results are ordered by where each control enters, starting from the in-house model on which the controls were first used (Section~\ref{sec:casea}) and ending with fine-tuning and the readout (Sections~\ref{sec:fullft} and~\ref{sec:readout}).

\subsection{Developing the in-house model}
\label{sec:casea}

The values in this subsection are the development record of the in-house model, copied from its project log because the original probe outputs were not kept, and are not regenerated (Appendix~\ref{app:devlog}).
Each architectural change was kept or dropped on the probe.
A learned-query channel mixer was dropped when its encoder probe on Lee2019 fell to \numDevVfiveEncProbe\ against \numDevRawProbe\ for the raw signal.
Its query weights had barely moved, their standard deviation going from \numDevQStdStart\ to \numDevQStdEnd\ over \numDevQSteps\ steps, and the run was stopped at \numDevVfiveStop.
Its replacement, per-channel tokens, kept the signal the raw probe sees (\numDevVeightEncProbe\ against \numDevRawProbe).
A plain convolutional patch embedder then lowered the probe to \numDevConvProbe.
In turn, the sinc front-end that replaced it raised the probe to \numDevSincProbe, above the \numDevSincBp\ of \arm{bp} in the same diagnostic, although Section~\ref{sec:inhouse} shows that its filters are low-pass rather than band-pass.

In the final run the pretrained-minus-random probe gain peaked at different checkpoints per benchmark.
It peaked at \numDevTrajApavaStep\ steps on APAVA (\numDevTrajApavaPeak\ AUROC), but at \numDevTrajBciStep\ on BCI IV 2a (\numDevTrajBciPeak\ accuracy) and \numDevTrajLeeStep\ on Lee2019 (\numDevTrajLeePeak\ accuracy).
Then, by 40{,}000 steps, the gain had fallen on three of four benchmarks (\numDevTrajApavaEnd, \numDevTrajBciEnd, \numDevTrajTuabEnd\ and \numDevTrajLeeEnd), so the run was stopped and the APAVA-peak checkpoint kept.
That kept checkpoint still does not beat \arm{bp} on most cells, which is the observation that turned the project into the evaluation reported next (Table~\ref{tab:classical}).

\subsection{Choosing the layer on test inflates layer-wise results}
\label{sec:selection}

\begin{figure}[t]\centering
\includegraphics[width=\textwidth]{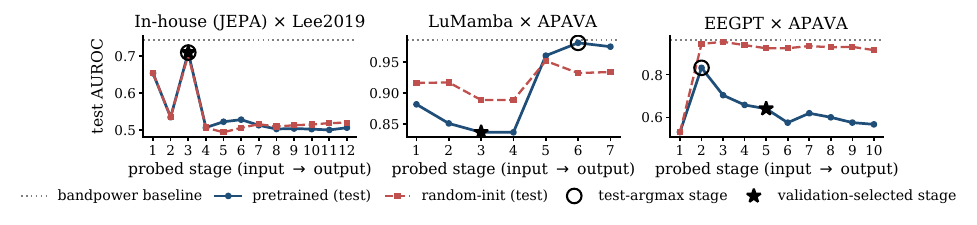}
\caption{Layer-wise frozen probe (test AUROC) for three cells of the leakage-free benchmarks (all cells in Appendix~\ref{app:curves}). Solid: \arm{pre}; dashed: \arm{rand}; dotted: \arm{bp}; circle: test-argmax stage; star: validation-selected stage. Left: a genuine early-stage peak that the random twin reproduces. Middle: validation picks an early stage while the test argmax sits one stage before the output. Right: the largest selection penalty.}
\label{fig:layerwise}
\end{figure}

Re-selecting the probed stage on validation lowers the reported peak by \numSelPenaltyMean\ on average and by up to \numSelPenaltyMax\ in one cell (Table~\ref{tab:valsel}).
These \numSelCells\ cells cover the three leakage-free benchmarks, six models under both initialisations.
These include \numSelFlip\ cells in which the peak-minus-output gap is positive under test selection and below $-$\numSelFlipThr\ under validation selection.
In those cells, therefore, the reading that an intermediate layer beats the output was an artefact of choosing the layer on test.
LuMamba on APAVA shows the pattern: its test-selected gap of \numSelLumambaApavaGapTest\ becomes \numSelLumambaApavaGapVal\ once validation picks the stage (Figure~\ref{fig:layerwise}, middle).

Validation selection does not remove every intermediate peak.
A genuine gap survives in \numSelHonest\ cells, the largest for the in-house model on Lee2019 at \numSelOursLeeGapVal.
That cell's \arm{rand} curve follows the \arm{pre} one, so its gap belongs to the front-end architecture rather than to pretraining (Figure~\ref{fig:layerwise}, left).
A layer-wise probe therefore supports a claim about intermediate layers only when the layer is chosen on validation and the random twin is plotted beside it (Figure~\ref{fig:layerwise}).

\subsection{The band-power anchor is rarely beaten}
\label{sec:bandpower}

\begin{figure}[t]\centering
\includegraphics[width=\textwidth]{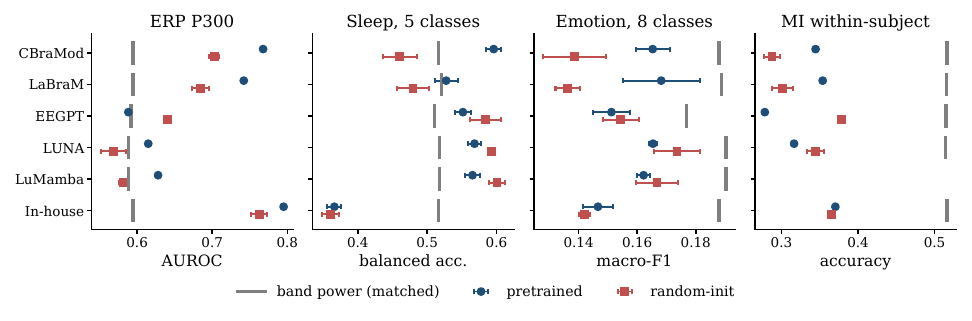}
\caption{Added task families, frozen probe at the validation-selected stage (mean $\pm$ std over \numCovSeeds\ seeds; a deterministic probe has no bar): \arm{pre} (circles), \arm{rand} (squares) and each model's matched \arm{bp} (grey bar). LuMamba was not run under the within-subject \acs{MI} protocol. Numbers in Appendix~\ref{app:coverage}.}
\label{fig:coverage}
\end{figure}

\begin{table}[t]\centering\small\setlength{\tabcolsep}{4pt}
\caption{Classical baselines on the original benchmarks, same splits: EEGNet trained from scratch (mean$\pm$std over 5 seeds), \acs{FBCSP} (motor imagery only), the band-power probe (range over the models' input pipelines) and the best pretrained model after full fine-tuning. AUROC except BCI IV 2a (accuracy, four classes).}
\label{tab:classical}
\begin{tabular}{lccccc}
\toprule
Method & APAVA & BCI IV 2a & Lee2019 & TUSL & TUAB \\
\midrule
EEGNet & 0.723{\scriptsize$\pm$0.197} & 0.482{\scriptsize$\pm$0.042} & 0.796{\scriptsize$\pm$0.046} & 0.535{\scriptsize$\pm$0.035} & 0.863{\scriptsize$\pm$0.009} \\
FBCSP & --- & 0.260 & 0.599 & --- & --- \\
Band power & 0.951--0.985 & 0.484--0.491 & 0.733--0.743 & 0.645--0.690 & 0.853--0.855 \\
Best model, fine-tuned & 0.959 {\scriptsize EEGPT} & 0.351 {\scriptsize CBraMod} & 0.848 {\scriptsize CBraMod} & 0.758 {\scriptsize CBraMod} & 0.874 {\scriptsize LaBraM} \\
\bottomrule
\end{tabular}
\end{table}

On emotion \arm{bp} beats all \numEmoNFm\ models at the validation-selected stage, against a chance level of \numEmoChance.
This gap is clear against an empirical null: retraining the \arm{bp} probe on shuffled training labels \numEmoNullK\ times per split seed gives \numEmoNullMeanMin--\numEmoNullMeanMax\ macro-F1 (95th percentile at most \numEmoNullQMax), while \arm{bp} scores \numEmoNullBpMin--\numEmoNullBpMax.
No shuffled fit reached the unshuffled score on any of the \numEmoNullSeeds\ seeds, so the permutation p-value is \numEmoNullP, the smallest that \numEmoNullK\ permutations can give.
On within-subject \ac{MI}, \arm{bp} beats all \numBciWNFm\ models while reaching up to \numBciWBpMax\ against a chance level of \numBciChance\ (Figure~\ref{fig:coverage}).
The same holds on the two \ac{MI} benchmarks of the original matrix, where no model exceeds \arm{bp} (\numMiAboveBp\ of \numMiCells\ cells, Appendix~\ref{app:orig}).
Switching \ac{MI} from the subject-independent to the within-subject protocol raises the validation-selected \arm{pre} peak by only \numBciWithinMinusCrossMin\ to \numBciWithinMinusCrossMax\ per model (Appendix~\ref{app:bci}).
Even the best peak under either protocol (\numBciWithinMaxPeak\ within, \numBciCrossMaxPeak\ across) stays below \arm{bp} under the same protocol (\numBciWBpMin--\numBciWBpMax\ within, \numOrigBciBpMin--\numOrigBciBpMax\ across).
Yet the shared pipeline leaves the within-subject \ac{MI} signal decodable, since EEGNet trained from scratch on the same split reaches \numBciWEegnet$\pm$\numBciWEegnetSD\ accuracy over \numBciWEegnetSeeds\ seeds, above \arm{bp}, and \ac{FBCSP} reaches \numBciWFbcsp.

Sleep staging is the closest case.
Here \numSleepAboveBp\ models clear \arm{bp}, \numSleepAboveBpClear\ of them by more than their seed standard deviation, but for \numSleepRandGePre\ of them the \arm{rand} twin scores higher still (Figure~\ref{fig:coverage}).
A second classical method is as hard to beat: EEGNet, a compact network trained from scratch \citep{lawhern2018eegnet}, reaches \numEegnetLee$\pm$\numEegnetLeeSD\ AUROC on Lee2019 and \numEegnetTuab$\pm$\numEegnetTuabSD\ on TUAB.
Both are within reach of the best fine-tuned foundation model (Table~\ref{tab:classical}).
Where the signal is spectral and the labels are few, the band-power anchor is the bar a pretrained representation has to clear first, and in this panel it often does not (Figure~\ref{fig:coverage}).

\subsection{A model that sees mostly slow components}
\label{sec:slow}

The front-end error also explains where the in-house model does well.
Under validation-selected frozen probes the in-house model ranks first among the evaluated models on \numRankSlowFirstOf\ motor-imagery and event-related tasks (Table~\ref{tab:valsel}, Figure~\ref{fig:coverage}).
Its margin over the next model is \numRankOursLee\ against \numRankNextLee\ on Lee2019 and \numRankOursErp\ against \numRankNextErp\ on P300.
It ranks last on \numRankSpecLastOf\ tasks whose classical markers are band powers, APAVA, sleep staging and emotion \citep{babiloni2009ad}.
This split is consistent with what the front-end passes, since slow and event-locked potentials survive a sub-hertz low-pass filter far better than band power above a few hertz (Section~\ref{sec:inhouse}).

On motor imagery it also suggests that the frozen representations of the five public models expose little of the mu and beta power that \arm{bp} reads (Section~\ref{sec:bandpower}).
A model that sees almost none of that power outranks them, while \arm{bp} outranks every model (Table~\ref{tab:valsel}).
The comparison holds for frozen representations only, since the in-house model was developed against Lee2019 (Section~\ref{sec:limitations}).
Full fine-tuning also reverses the order, with the \arm{pre} CBraMod at \numFtPreCbramodLee\ on Lee2019 against \numFtPreOursLee\ for the in-house model.

\subsection{Random initialisation often matches pretraining}
\label{sec:random}

On the added tasks \arm{rand} equals or beats \arm{pre} for \numSleepRandGePre\ of \numSleepNFm\ models on sleep, \numEmoRandGePre\ of \numEmoNFm\ on emotion, \numBciWRandGePre\ of \numBciWNFm\ on within-subject \ac{MI}, and \numErpRandGePre\ of \numErpNFm\ on \ac{ERP} (Figure~\ref{fig:coverage}).
In the other direction, the \arm{pre} probe on \ac{ERP} detection sits above its \arm{rand} twin for \numErpPreAboveRand\ of the \numErpNFm\ models, by \numErpGapMinPos\ to \numErpGapMaxPos.
Within sleep staging, the one large frozen-probe gain in the panel is CBraMod's \numSleepCbramodGap, while LUNA, EEGPT and LuMamba score higher at random initialisation than pretrained.

Two checks bound how far this reading goes.
The random twin is itself a sample: two independent validation-selected draws of the same \arm{rand} cell differ by \numRandDrawDiffMean\ on average (Appendix~\ref{app:randdraw}).
The largest difference between \arm{rand} draws is \numRandDrawDiffMax, for \numRandDrawMaxCell\ (\numRandDrawMaxA\ against \numRandDrawMaxB), so a single draw bounds a frozen gain only to that precision.
A second check cuts the label budget to \numBudgetLowFrac\% of the training set and finds no low-label advantage (Appendix~\ref{app:budget}).
A gap that grows by more than \numBudgetGrowThr\ over its full-budget value occurs in \numBudgetGrowMean\ of \numBudgetCells\ cells on the seed mean, and only \numBudgetGrowMeanPos\ of those are positive at the low budget.
Under a frozen probe, then, much of what reads as a pretrained representation is already present at random initialisation, which is why the twin has to be reported with every cell (Figure~\ref{fig:coverage}).

\subsection{Fine-tuning gains mostly sit within seed noise}
\label{sec:fullft}

\begin{figure}[t]\centering
\includegraphics[width=\textwidth]{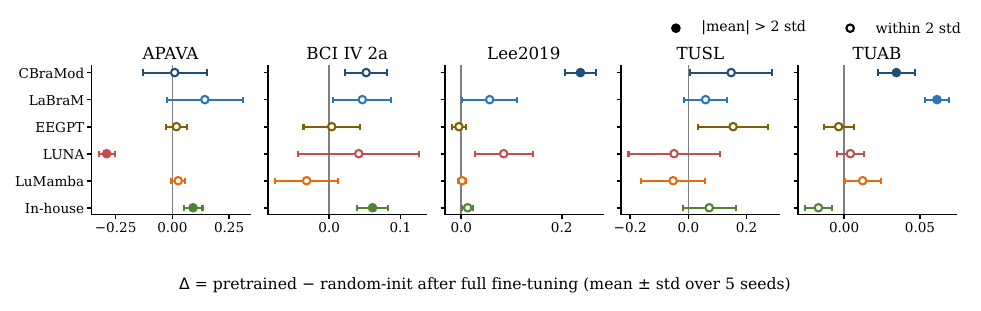}
\caption{Full fine-tuning gain $\Delta$ per model and benchmark. Filled markers: mean beyond twice the seed standard deviation. The in-house Lee2019 cell carries developer contact with that benchmark (Section~\ref{sec:limitations}).}
\label{fig:fullft}
\end{figure}

\begin{table}[t]\centering\small\setlength{\tabcolsep}{3.5pt}
\caption{Gain $\Delta$ of full fine-tuning, pretrained minus random-init at the best-validation epoch (test metric), mean$\pm$population std over 5 seeds, on the five original benchmarks; bold = mean beyond twice its seed std. $k$ = datasets with a positive mean. The in-house Lee2019 cells carry developer contact with that benchmark (Section~\ref{sec:limitations}).}
\label{tab:fullft}
\begin{tabular}{lcccccc}
\toprule
Model & APAVA & BCI IV 2a & Lee2019 & TUSL & TUAB & mean ($k$/5) \\
\midrule
CBraMod & +0.010{\scriptsize$\pm$0.141} & +0.052{\scriptsize$\pm$0.030} & \textbf{+0.237{\scriptsize$\pm$0.030}} & +0.147{\scriptsize$\pm$0.141} & \textbf{+0.034{\scriptsize$\pm$0.012}} & +0.096 (5) \\
LaBraM & +0.143{\scriptsize$\pm$0.166} & +0.047{\scriptsize$\pm$0.041} & +0.057{\scriptsize$\pm$0.054} & +0.059{\scriptsize$\pm$0.075} & \textbf{+0.061{\scriptsize$\pm$0.008}} & +0.073 (5) \\
EEGPT & +0.018{\scriptsize$\pm$0.045} & +0.003{\scriptsize$\pm$0.040} & $-$0.005{\scriptsize$\pm$0.014} & +0.154{\scriptsize$\pm$0.119} & $-$0.004{\scriptsize$\pm$0.010} & +0.033 (3) \\
LUNA & \textbf{$-$0.290{\scriptsize$\pm$0.035}} & +0.041{\scriptsize$\pm$0.085} & +0.084{\scriptsize$\pm$0.058} & $-$0.049{\scriptsize$\pm$0.157} & +0.004{\scriptsize$\pm$0.009} & $-$0.042 (3) \\
LuMamba & +0.026{\scriptsize$\pm$0.030} & $-$0.032{\scriptsize$\pm$0.044} & +0.002{\scriptsize$\pm$0.008} & $-$0.052{\scriptsize$\pm$0.110} & +0.012{\scriptsize$\pm$0.012} & $-$0.009 (3) \\
In-house (JEPA) & \textbf{+0.091{\scriptsize$\pm$0.042}} & \textbf{+0.061{\scriptsize$\pm$0.022}} & +0.013{\scriptsize$\pm$0.010} & +0.072{\scriptsize$\pm$0.091} & $-$0.017{\scriptsize$\pm$0.009} & +0.044 (4) \\
\bottomrule
\end{tabular}
\end{table}

Of the \numFtCells\ pairs, \numFtPos\ have a positive mean gain after full fine-tuning, but only \numFtPosTwoSd\ positive pairs are clear of twice their seed standard deviation and \numFtNegTwoSd\ negative pair is (Figure~\ref{fig:fullft}, Table~\ref{tab:fullft}).
Paired over seeds, the 95\% interval of the gain excludes zero in \numFtPairCiPos\ positive and \numFtPairCiNeg\ negative of the \numFtPairCells\ cells.
After Benjamini--Hochberg correction across all cells \citep{benjamini1995fdr}, \numFtPairBh\ cells remain, \numFtPairBhTuh\ of them on the TUH-derived benchmarks that overlap most pretraining corpora.
These gains are also easy to misread from one seed, as LaBraM on TUSL gives \numFtLabramTuslSeedOne\ at one seed but \numFtLabramTuslDelta$\pm$\numFtLabramTuslDeltaSD\ over \numFtSeeds, and CBraMod on BCI IV 2a moves from \numFtCbramodBciSeedOne\ to \numFtCbramodBciDelta$\pm$\numFtCbramodBciDeltaSD.
Fine-tuning therefore buys a clear gain only in isolated cells, spread over \numFtPosTwoSdModels\ models with no model clear on more than \numFtPosTwoSdMaxPerModel\ of the five original benchmarks (Figure~\ref{fig:fullft}).

The clear negative pair, LUNA on APAVA at \numFtLunaApavaDelta$\pm$\numFtLunaApavaDeltaSD, persists under three shared learning rates (\numTfiveDeltaLrLow, \numTfiveDeltaLrMid\ and \numTfiveDeltaLrHigh, \numTfiveSeeds\ seeds each), so the asymmetric schedule did not manufacture it (Appendix~\ref{app:lr}).
A fresh asymmetric run of that pair gives \numTfiveDeltaAsym, within seed noise of the matrix value.
Its frozen-probe gain is fragile under validation selection: \numBudgetLunaApavaFullGap\ or \numSelLunaApavaValGap\ in two runs, because near-tied validation values pick different stages, and \numFtLunaApavaProbeDelta\ test-selected.
At the full label budget, though, every reading finds the \arm{pre} representation no worse than its \arm{rand} twin, so the loss appears only once the weights are trained.
That frozen-positive, fine-tuning-negative label is the only one of the four modes stable across thresholds (\numModeFrozenOnlyStable\ pairs at every threshold, while both-positive moves between \numModeBothPosLo\ and \numModeBothPosHi; Appendix~\ref{app:fourmode}).

All of these gains were measured under one shared schedule.
Tuning the learning rate per model on validation would have added \numTfourPenMean\ on average (range \numTfourPenMin\ to \numTfourPenMax\ over \numTfourCells\ pairs at one seed, Appendix~\ref{app:tfour}).
The largest penalty falls on the in-house model on APAVA, so the canonical numbers are a modest lower bound.

\subsection{Nonlinear readouts recover little}
\label{sec:readout}

Replacing mean pooling by a trainable attention-pooling layer in the style of the models' own heads does not recover the output layer (Appendix~\ref{app:readout}).
Run for two models on the three leakage-free benchmarks, its score minus the mean-pool probe lies between \numAttnDMin\ and \numAttnDMax\ in all \numAttnCells\ cells.
The other readout, a one-hidden-layer \ac{MLP} probe, lifts the output stage by more than three points in at most \numMlpRecoverMax\ of \numMlpCells\ cells at any of \numMlpSeeds\ seeds (Appendix~\ref{app:readout}).
The losses measured by the linear probe are thus not an artefact of mean pooling or of linearity (Appendix~\ref{app:readout}).


\section{Discussion}
\label{sec:discussion}

\subsection{Reporting practices}
\label{sec:practices}

The results suggest four habits for anyone who evaluates an EEG foundation model.
First, report the matched \arm{rand} encoder and \arm{bp} on the same split for every cell (Section~\ref{sec:random}).
Without the \arm{rand} twin \numSleepRandGePre\ of \numSleepNFm\ sleep cells would read as pretraining gains, and without \arm{bp} \numEmoBelowBp\ of \numEmoNFm\ emotion cells would read as usable representations.
Second, select the probed layer on validation and publish the full layer-wise curve, because test selection added \numSelPenaltyMean\ on average here (Section~\ref{sec:selection}).
Third, report seeds together with what they vary, since a fixed split turns repeated frozen probes into replicates and a single seed can reverse the sign of a fine-tuning gain (Section~\ref{sec:fullft}).
Fourth, present shared-protocol numbers as the floor of a fair comparison, not as a ranking of architectures, and report the tuning penalty where it can be measured \citep{musgrave2020metric,pineau2018reproducibility}.
None of the four needs new data or large compute, because \arm{rand}, \arm{bp} and the validation split are already part of any evaluation that reports a pretrained model.

\subsection{Scope of the conclusions}
\label{sec:notestablished}

The results do not establish a causal ranking of pretraining objectives or input-handling designs, because in a panel of six those choices are confounded with corpus, tuning and implementation (Appendix~\ref{app:typology}).
Nor do they cover every EEG paradigm, since the added tasks enter in reduced single-window form and clinical event detection and sleep sequence models lie outside the protocol (Section~\ref{sec:setup}).
Finally, they do not show that the diagnostic protocol helps developers of other models, since it was used during development on one model only (Section~\ref{sec:diagnostic}).

\subsection{Limitations}
\label{sec:limitations}

The evaluation covers five public checkpoints and the in-house model, one linear-probe family with two nonlinear checks, one fine-tuning recipe and five-second single windows (Section~\ref{sec:protocols}).
The original matrix also overlaps with pretraining data, since five of the six models and the in-house corpus contain TUH-derived recordings \citep{obeid2016tuh}.
TUSL and TUAB may therefore have been seen in pretraining, which is why the added task families were chosen outside TUH (Section~\ref{sec:setup}).
That released matrix also still uses test-selected stages in two places, the frozen probe on TUSL and TUAB and the layer-wise re-probe after fine-tuning, which we cite only as upper bounds (Appendix~\ref{app:orig}).
The in-house front-end did not implement its design, and we report the model as released rather than retrain it with corrected filters, so the in-house rows describe a model that sees mostly sub-hertz content (Section~\ref{sec:inhouse}).

A limit of its own comes from how the protocol was produced.
The probe protocol was fixed while diagnosing the in-house model, whose iterations were judged on held-out pretraining windows and on Lee2019.
The in-house Lee2019 cells therefore carry developer contact, and the validation rule and the added tasks came after the first results were seen.
This makes the study exploratory rather than pre-registered, and its conclusions are scoped to the tested panel and protocols \citep{pineau2018reproducibility}.


\section{Conclusion}
\label{sec:conclusion}

With a matched random twin, a matched band-power baseline and a validation-selected layer in place, most of the gains that EEG foundation models report under their own protocols shrink or vanish in this panel.
Fine-tuning gains mostly sit within seed noise, and the one clearly negative pair survives matched learning rates.
The four practices cost little, and the per-cell results are released\releaseurl\ so that each claim here can be checked against its own numbers.
The natural next step is to run the same controls during pretraining, where they can still change a design.

\bibliography{references}
\bibliographystyle{tmlr}

\appendix

\section{Models and checkpoints}
\label{app:models}

\begin{table}[H]\centering\small
\caption{Model checkpoints: repository, file and the revision pinned when the paper is built (hf.lock.json).}
\label{tab:models}
\resizebox{\textwidth}{!}{%
\begin{tabular}{llll}
\toprule
Model & Hugging Face repository & File & Revision \\
\midrule
EEG-OmniJEPA (ours, REVE 25k) & not released & \texttt{ckpts/step\_025000\_weights\_only.pt} & \texttt{9567ab6053} \\
LaBraM-Base & \texttt{braindecode/labram-pretrained} & \texttt{model.safetensors} & \texttt{0563b6c626} \\
CBraMod & \texttt{braindecode/cbramod-pretrained} & \texttt{model.safetensors} & \texttt{584cdc4159} \\
LUNA-Base & \texttt{PulpBio/LUNA} & \texttt{LUNA\_base.safetensors} & \texttt{999c1af0fd} \\
LuMamba LeJEPA-only & \texttt{PulpBio/LuMamba} & \texttt{LuMamba\_LeJEPAOnly\_128slices.safetensors} & \texttt{fd3da57e34} \\
EEGPT-Base & \texttt{braindecode/eegpt-pretrained} & \texttt{model.safetensors} & \texttt{e41cb3ae2c} \\
\bottomrule
\end{tabular}}
\end{table}

\begin{table}[H]\centering\small\setlength{\tabcolsep}{4pt}
\caption{Input each public model was pretrained on (its paper or official release) and the input its adapter feeds in this audit, after the shared chain of Table~\ref{tab:datasets}. Sampling rate in Hz, window in seconds.}
\label{tab:native}
\begin{tabular}{lcccc}
\toprule
Model & pretrained rate & fed rate & pretrained window & fed window \\
\midrule
CBraMod & 200 & 200 & 30 & 5 \\
LaBraM & 200 & 200 & $\le$256 one-second patches & 15 \\
EEGPT & 256 & 250 & 4 & 4 \\
LUNA & 256 & 256 & 5 & 5 \\
LuMamba & 256 & 256 & 5 & 5 \\
\bottomrule
\end{tabular}
\end{table}

Table~\ref{tab:models} gives each model's checkpoint as a repository, file and revision, read from the evaluation code and pinned when the paper is built.

\section{Protocol constants}
\label{app:constants}

\begin{table}[H]\centering\small
\caption{Protocol constants, read from the evaluation code at build time.}
\label{tab:constants}
\begin{tabular}{lp{10.5cm}}
\toprule
Item & Value \\
\midrule
Sampling rate / window & 200\,Hz, five seconds, common-average reference, band-pass and notch filtering \\
Bandpower baseline & log band power in five bands (0.5--4, 4--8, 8--13, 13--30, 30--75\,Hz) per channel, then the same probe \\
Linear probe & mean-pool over tokens, standardise on the training split, L2 logistic regression ($C=1.0$, at most 3000 iterations) \\
Stage selection & argmax of the validation metric over stages; test metric reported at that stage \\
Full fine-tuning & each model's own head; learning rate 5e-5 (pretrained) / 5e-4 (random-init); up to 50 epochs, batch 32; best-validation epoch reported \\
Seeds & 5 (original benchmarks) or 3 (added tasks); population std throughout \\
\bottomrule
\end{tabular}
\end{table}

Table~\ref{tab:constants} lists the preprocessing, probe and fine-tuning constants, read from the evaluation code when the paper is built.

\section{The in-house front-end as implemented}
\label{app:frontend}

\begin{table}[H]\centering\small
\caption{What the released in-house front-end passes. Top: share of spectral power per band (\%) in Lee2019 (first 32 windows, all channels), at the input, at the output of the released front-end, and at the output of the same filters with the time index in samples ($n$ instead of $n/f_s$ in Eq.~\ref{eq:frontend}; not retrained). Bottom: gain (output over input amplitude) for pure sines, for four of the filters.}
\label{tab:frontend}
\begin{tabular}{lcccccc}
\toprule
Signal & 0--1\,Hz & 1--4\,Hz & 4--8\,Hz & 8--13\,Hz & 13--30\,Hz & 30--75\,Hz \\
\midrule
Input & 37.4 & 29.0 & 6.3 & 7.5 & 6.9 & 12.7 \\
Front-end as released & 91.2 & 7.2 & 0.8 & 0.3 & 0.3 & 0.1 \\
Same filters, $n$ in samples & 0.2 & 9.4 & 16.0 & 20.4 & 18.6 & 34.9 \\
\bottomrule
\end{tabular}

\medskip
\begin{tabular}{lccccccc}
\toprule
Filter (designed band) & 0.3\,Hz & 2\,Hz & 6\,Hz & 10\,Hz & 20\,Hz & 40\,Hz & 70\,Hz \\
\midrule
0 (2--5\,Hz), as released & 169.71 & 0.06 & 0.08 & 0.08 & 0.08 & 0.08 & 0.08 \\
0 (2--5\,Hz), $n$ in samples & 0.06 & 15.90 & 0.45 & 0.00 & 0.00 & 0.00 & 0.00 \\
5 (10--13\,Hz), as released & 169.18 & 0.05 & 0.07 & 0.07 & 0.07 & 0.07 & 0.07 \\
5 (10--13\,Hz), $n$ in samples & 0.01 & 0.02 & 0.03 & 27.63 & 0.00 & 0.00 & 0.00 \\
24 (38--42\,Hz), as released & 161.70 & 0.06 & 0.03 & 0.01 & 0.00 & 0.00 & 0.00 \\
24 (38--42\,Hz), $n$ in samples & 0.01 & 0.01 & 0.01 & 0.01 & 0.01 & 28.42 & 0.00 \\
63 (97--100\,Hz), as released & 112.06 & 0.02 & 0.01 & 0.04 & 0.05 & 0.06 & 0.06 \\
63 (97--100\,Hz), $n$ in samples & 0.00 & 0.00 & 0.00 & 0.00 & 0.00 & 0.00 & 0.00 \\
\bottomrule
\end{tabular}
\end{table}

Table~\ref{tab:frontend} gives the spectral power that the released front-end passes on Lee2019 and the gains of four of its filters for pure sines.
Beside each it shows what those filters pass once the scaling error is corrected, as measured by the script released with the paper.

\section{Layer-wise curves and validation selection}
\label{app:curves}

\begin{figure}[H]\centering
\includegraphics[width=\textwidth]{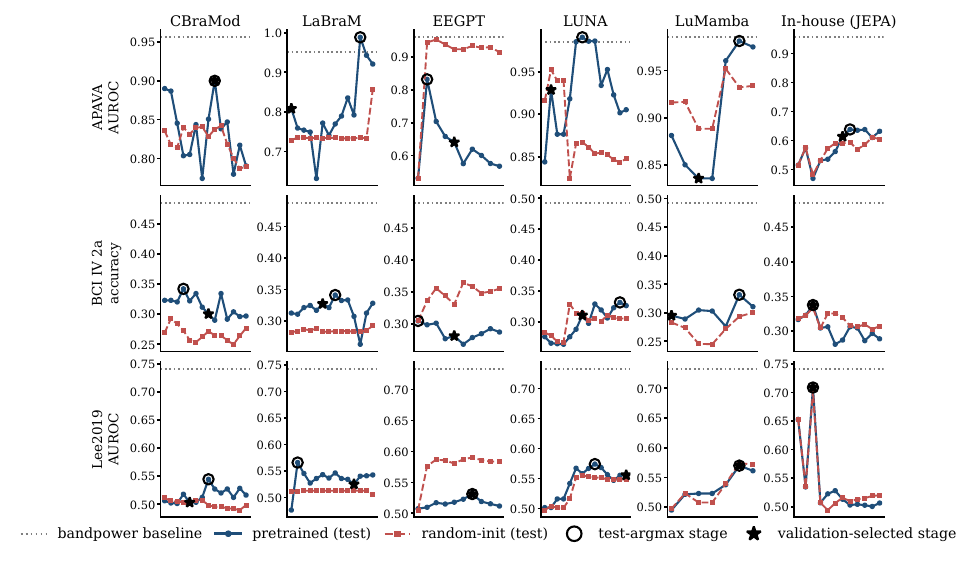}
\caption{Layer-wise frozen linear probe for every \arm{pre} cell of the three leakage-free benchmarks (rows: datasets; columns: models; x-axis: probed stage from input to output). Solid: \arm{pre}; dashed: \arm{rand}; dotted: \arm{bp}; circle: test-argmax stage; star: validation-selected stage.}
\label{fig:layerwise-all}
\end{figure}

\begin{table}[H]\centering\footnotesize\setlength{\tabcolsep}{3pt}
\caption{Frozen probe on the three leakage-free original benchmarks, both initialisations (36 cells): validation-selected peak minus output stage (selection penalty = test-argmax minus validation-selected peak, in parentheses). APAVA and Lee2019 report AUROC, BCI IV 2a accuracy; the in-house Lee2019 cells carry developer contact with that benchmark.}
\label{tab:valsel}
\begin{tabular}{lcccccc}
\toprule
Model & \multicolumn{3}{c}{pretrained} & \multicolumn{3}{c}{random-init} \\
\cmidrule(lr){2-4}\cmidrule(lr){5-7}
 & APAVA & BCI IV 2a & Lee2019 & APAVA & BCI IV 2a & Lee2019 \\
\midrule
CBraMod & +0.110 {\scriptsize(+0.000)} & +0.003 {\scriptsize(+0.042)} & $-$0.013 {\scriptsize(+0.041)} & +0.011 {\scriptsize(+0.041)} & $-$0.004 {\scriptsize(+0.020)} & +0.006 {\scriptsize(+0.007)} \\
LaBraM & $-$0.113 {\scriptsize(+0.181)} & $-$0.001 {\scriptsize(+0.014)} & $-$0.018 {\scriptsize(+0.041)} & +0.000 {\scriptsize(+0.000)} & $-$0.010 {\scriptsize(+0.010)} & +0.007 {\scriptsize(+0.000)} \\
EEGPT & +0.073 {\scriptsize(+0.191)} & $-$0.006 {\scriptsize(+0.023)} & +0.019 {\scriptsize(+0.000)} & +0.015 {\scriptsize(+0.024)} & $-$0.007 {\scriptsize(+0.016)} & $-$0.008 {\scriptsize(+0.014)} \\
LUNA & +0.023 {\scriptsize(+0.062)} & $-$0.016 {\scriptsize(+0.021)} & +0.000 {\scriptsize(+0.018)} & $-$0.023 {\scriptsize(+0.128)} & +0.023 {\scriptsize(+0.000)} & $-$0.000 {\scriptsize(+0.007)} \\
LuMamba & $-$0.139 {\scriptsize(+0.145)} & $-$0.016 {\scriptsize(+0.036)} & +0.009 {\scriptsize(+0.000)} & $-$0.002 {\scriptsize(+0.020)} & $-$0.007 {\scriptsize(+0.007)} & +0.000 {\scriptsize(+0.002)} \\
In-house (JEPA) & $-$0.018 {\scriptsize(+0.025)} & +0.049 {\scriptsize(+0.000)} & +0.202 {\scriptsize(+0.000)} & $-$0.074 {\scriptsize(+0.080)} & +0.026 {\scriptsize(+0.000)} & +0.183 {\scriptsize(+0.000)} \\
\bottomrule
\end{tabular}
\end{table}

Figure~\ref{fig:layerwise-all} extends Figure~\ref{fig:layerwise} to every pretrained cell of the three leakage-free benchmarks, and Table~\ref{tab:valsel} gives the validation-selected gaps and the selection penalties behind Section~\ref{sec:selection}.

\section{Added task families}
\label{app:coverage}

\begin{table}[t]\centering\small\setlength{\tabcolsep}{3.2pt}
\caption{Frozen linear probe on the four added task families: validation-selected stage, test metric, mean of 3 seeds; $\Delta$ = pretrained $-$ random-init, computed before rounding; last row = five-band log-power baseline on the same split. Seed std $\le$ 0.024; on the within-subject protocols the split is deterministic, so pretrained is a point estimate and only random-init carries a seed band. The band-power baseline is computed on each model's own preprocessed input; each model is compared with its own baseline (seed mean, std $\le$ 0.023), and the last row gives the range over models. \acs{MI} within-subject: 5 models, as LuMamba was not run under that protocol.}
\label{tab:coverage}
\begin{tabular}{lcccccccc}
\toprule
Model & \multicolumn{2}{c}{ERP P300 (AUROC)} & \multicolumn{2}{c}{Sleep 5-cl. (bal.\ acc.)} & \multicolumn{2}{c}{Emotion 8-cl. (macro-F1)} & \multicolumn{2}{c}{MI within-subj. (acc.)} \\
 & pre / rand & $\Delta$ & pre / rand & $\Delta$ & pre / rand & $\Delta$ & pre / rand & $\Delta$ \\
\midrule
CBraMod & 0.768 / 0.703 & +0.065 & 0.596 / 0.461 & +0.135 & 0.165 / 0.139 & +0.027 & 0.344 / 0.288 & +0.057 \\
LaBraM & 0.742 / 0.684 & +0.058 & 0.527 / 0.479 & +0.048 & 0.168 / 0.136 & +0.032 & 0.354 / 0.301 & +0.052 \\
EEGPT & 0.588 / 0.641 & $-$0.052 & 0.552 / 0.584 & $-$0.033 & 0.151 / 0.154 & $-$0.003 & 0.278 / 0.379 & $-$0.100 \\
LUNA & 0.615 / 0.568 & +0.046 & 0.568 / 0.593 & $-$0.024 & 0.165 / 0.174 & $-$0.008 & 0.316 / 0.344 & $-$0.028 \\
LuMamba & 0.628 / 0.581 & +0.047 & 0.565 / 0.601 & $-$0.035 & 0.162 / 0.167 & $-$0.005 & --- & --- \\
In-house (JEPA) & 0.795 / 0.763 & +0.032 & 0.366 / 0.360 & +0.006 & 0.147 / 0.142 & +0.005 & 0.370 / 0.365 & +0.005 \\
\midrule
Bandpower & 0.589--0.594 &  & 0.511--0.520 &  & 0.177--0.190 &  & 0.514--0.516 &  \\
\bottomrule
\end{tabular}
\end{table}

\begin{table}[H]\centering\footnotesize\setlength{\tabcolsep}{3pt}
\caption{Table~\ref{tab:coverage} with seed standard deviations (3 seeds; a std of 0.000 marks a deterministic within-subject pretrained probe).}
\label{tab:coverage-sd}
\begin{tabular}{lcccccccc}
\toprule
Model & \multicolumn{2}{c}{ERP P300} & \multicolumn{2}{c}{Sleep 5-cl.} & \multicolumn{2}{c}{Emotion 8-cl.} & \multicolumn{2}{c}{MI within-subj.} \\
 & pretrained & random-init & pretrained & random-init & pretrained & random-init & pretrained & random-init \\
\midrule
CBraMod & 0.768{\scriptsize$\pm$0.000} & 0.703{\scriptsize$\pm$0.006} & 0.596{\scriptsize$\pm$0.011} & 0.461{\scriptsize$\pm$0.024} & 0.165{\scriptsize$\pm$0.006} & 0.139{\scriptsize$\pm$0.011} & 0.344{\scriptsize$\pm$0.000} & 0.288{\scriptsize$\pm$0.011} \\
LaBraM & 0.742{\scriptsize$\pm$0.000} & 0.684{\scriptsize$\pm$0.011} & 0.527{\scriptsize$\pm$0.017} & 0.479{\scriptsize$\pm$0.023} & 0.168{\scriptsize$\pm$0.013} & 0.136{\scriptsize$\pm$0.004} & 0.354{\scriptsize$\pm$0.000} & 0.301{\scriptsize$\pm$0.014} \\
EEGPT & 0.588{\scriptsize$\pm$0.000} & 0.641{\scriptsize$\pm$0.005} & 0.552{\scriptsize$\pm$0.012} & 0.584{\scriptsize$\pm$0.022} & 0.151{\scriptsize$\pm$0.006} & 0.154{\scriptsize$\pm$0.006} & 0.278{\scriptsize$\pm$0.000} & 0.379{\scriptsize$\pm$0.003} \\
LUNA & 0.615{\scriptsize$\pm$0.000} & 0.568{\scriptsize$\pm$0.017} & 0.568{\scriptsize$\pm$0.009} & 0.593{\scriptsize$\pm$0.004} & 0.165{\scriptsize$\pm$0.001} & 0.174{\scriptsize$\pm$0.008} & 0.316{\scriptsize$\pm$0.000} & 0.344{\scriptsize$\pm$0.011} \\
LuMamba & 0.628{\scriptsize$\pm$0.000} & 0.581{\scriptsize$\pm$0.005} & 0.565{\scriptsize$\pm$0.011} & 0.601{\scriptsize$\pm$0.011} & 0.162{\scriptsize$\pm$0.002} & 0.167{\scriptsize$\pm$0.007} & --- & --- \\
In-house (JEPA) & 0.795{\scriptsize$\pm$0.000} & 0.763{\scriptsize$\pm$0.011} & 0.366{\scriptsize$\pm$0.010} & 0.360{\scriptsize$\pm$0.012} & 0.147{\scriptsize$\pm$0.005} & 0.142{\scriptsize$\pm$0.002} & 0.370{\scriptsize$\pm$0.000} & 0.365{\scriptsize$\pm$0.001} \\
\midrule
Bandpower & \multicolumn{2}{c}{0.589--0.594} & \multicolumn{2}{c}{0.511--0.520} & \multicolumn{2}{c}{0.177--0.190} & \multicolumn{2}{c}{0.514--0.516} \\
\bottomrule
\end{tabular}
\end{table}

Tables~\ref{tab:coverage} and~\ref{tab:coverage-sd} give the numbers behind Figure~\ref{fig:coverage}, with seed standard deviations.

\section{Random-initialisation draws}
\label{app:randdraw}

The validation-selected run builds the frozen-probe random encoder under a fixed seed, and the label-budget re-extraction builds it without one.
These \numRandDrawCells\ \arm{rand} cells, probed with the same probe seed, are therefore two independent draws.
Their mean absolute difference of \numRandDrawDiffMean\ is thus the resolution of any single frozen \arm{rand} value in this paper.

\section{Readout robustness}
\label{app:readout}

\begin{table}[H]\centering\small
\caption{Trainable attention-pooling readout on frozen features (3 seeds) against the mean-pool linear probe at the output stage. $d$ = attention minus mean-pool; a non-positive $d$ means the native-style readout does not recover the output layer.}
\label{tab:attn}
\begin{tabular}{llccc}
\toprule
Model & Dataset & mean-pool final & attention final & $d$ \\
\midrule
LUNA & APAVA & 0.905 & 0.886{\scriptsize$\pm$0.035} & $-$0.019 \\
LUNA & BCI IV 2a & 0.326 & 0.309{\scriptsize$\pm$0.003} & $-$0.017 \\
LUNA & Lee2019 & 0.556 & 0.523{\scriptsize$\pm$0.009} & $-$0.033 \\
In-house (JEPA) & APAVA & 0.631 & 0.562{\scriptsize$\pm$0.004} & $-$0.069 \\
In-house (JEPA) & BCI IV 2a & 0.286 & 0.275{\scriptsize$\pm$0.017} & $-$0.011 \\
In-house (JEPA) & Lee2019 & 0.506 & 0.499{\scriptsize$\pm$0.004} & $-$0.007 \\
\bottomrule
\end{tabular}
\end{table}

\begin{table}[H]\centering\small
\caption{One-hidden-layer \acs{MLP} probe minus linear probe at the output stage, pretrained encoders, mean$\pm$std over 3 seeds. Positive values would indicate non-linearly re-encoded information.}
\label{tab:mlp}
\begin{tabular}{lccc}
\toprule
Model & APAVA & BCI IV 2a & Lee2019 \\
\midrule
CBraMod & $-$0.017{\scriptsize$\pm$0.021} & $-$0.015{\scriptsize$\pm$0.006} & +0.004{\scriptsize$\pm$0.001} \\
LaBraM & +0.039{\scriptsize$\pm$0.013} & $-$0.005{\scriptsize$\pm$0.018} & $-$0.021{\scriptsize$\pm$0.001} \\
EEGPT & $-$0.132{\scriptsize$\pm$0.009} & $-$0.019{\scriptsize$\pm$0.012} & $-$0.003{\scriptsize$\pm$0.004} \\
LUNA & $-$0.044{\scriptsize$\pm$0.006} & $-$0.019{\scriptsize$\pm$0.019} & $-$0.026{\scriptsize$\pm$0.002} \\
LuMamba & $-$0.044{\scriptsize$\pm$0.002} & $-$0.007{\scriptsize$\pm$0.009} & $-$0.026{\scriptsize$\pm$0.002} \\
In-house (JEPA) & $-$0.036{\scriptsize$\pm$0.006} & $-$0.026{\scriptsize$\pm$0.002} & $-$0.001{\scriptsize$\pm$0.004} \\
\bottomrule
\end{tabular}
\end{table}

Tables~\ref{tab:attn} and~\ref{tab:mlp} give the attention-pooling and \ac{MLP} readouts behind Section~\ref{sec:readout}.

\section{Tuned anchor for fine-tuning}
\label{app:tfour}

\begin{table}[t]\centering\small
\caption{Tuned anchor for full fine-tuning (one seed): each model's learning rate chosen from 2e-5, 5e-5, 1e-4 by minimum validation loss, against the canonical rate of the main matrix. The penalty is what the canonical protocol gives away; a positive value means tuning would have helped.}
\label{tab:tfour}
\begin{tabular}{llccc c}
\toprule
Model & Dataset & chosen LR & tuned & canonical & penalty \\
\midrule
CBraMod & APAVA & 5e-5 & 0.883 & 0.883 & +0.000 \\
 & BCI IV 2a & 2e-5 & 0.440 & 0.363 & +0.077 \\
 & Lee2019 & 1e-4 & 0.837 & 0.840 & $-$0.003 \\
\addlinespace[2pt]
LaBraM & APAVA & 2e-5 & 0.932 & 0.951 & $-$0.019 \\
 & BCI IV 2a & 2e-5 & 0.319 & 0.309 & +0.010 \\
 & Lee2019 & 1e-4 & 0.634 & 0.578 & +0.055 \\
\addlinespace[2pt]
EEGPT & APAVA & 5e-5 & 0.953 & 0.953 & +0.000 \\
 & BCI IV 2a & 5e-5 & 0.331 & 0.331 & +0.000 \\
 & Lee2019 & 1e-4 & 0.503 & 0.495 & +0.008 \\
\addlinespace[2pt]
LUNA & APAVA & 2e-5 & 0.805 & 0.724 & +0.081 \\
 & BCI IV 2a & 5e-5 & 0.336 & 0.336 & +0.000 \\
 & Lee2019 & 1e-4 & 0.631 & 0.584 & +0.048 \\
\addlinespace[2pt]
In-house (JEPA) & APAVA & 1e-4 & 0.574 & 0.426 & +0.148 \\
 & BCI IV 2a & 2e-5 & 0.295 & 0.240 & +0.055 \\
 & Lee2019 & 1e-4 & 0.514 & 0.505 & +0.008 \\
\bottomrule
\end{tabular}
\end{table}

Table~\ref{tab:tfour} gives the per-model learning-rate grid, in which the canonical rate was kept in \numTfourCanonKept\ pairs and tuning lowered the test score in \numTfourNeg.

\section{Matched learning rates}
\label{app:lr}

\begin{table}[H]\centering\small
\caption{LUNA on APAVA under full fine-tuning, re-run on one machine: the asymmetric schedule of the main matrix and three shared learning rates (AUROC at the best-validation epoch, mean$\pm$std over five seeds). The asymmetric row is a fresh run and differs from the Table~\ref{tab:fullft} value within seed noise.}
\label{tab:lr}
\begin{tabular}{lccc}
\toprule
Learning rate (pretrained / random-init) & pretrained & random-init & $\Delta$ \\
\midrule
5e-5 / 5e-4 (asymmetric, main matrix) & 0.663{\scriptsize$\pm$0.052} & 0.935{\scriptsize$\pm$0.044} & $-$0.272 \\
2e-5 / 2e-5 & 0.763{\scriptsize$\pm$0.078} & 0.947{\scriptsize$\pm$0.023} & $-$0.184 \\
5e-5 / 5e-5 & 0.663{\scriptsize$\pm$0.052} & 0.950{\scriptsize$\pm$0.027} & $-$0.287 \\
1e-4 / 1e-4 & 0.824{\scriptsize$\pm$0.059} & 0.948{\scriptsize$\pm$0.028} & $-$0.124 \\
\bottomrule
\end{tabular}
\end{table}

Table~\ref{tab:lr} gives the matched-learning-rate runs of LUNA on APAVA, whose gain stays negative at every shared rate.

\section{Label budget}
\label{app:budget}

\begin{table}[H]\centering\small\setlength{\tabcolsep}{4pt}
\caption{Pretrained minus random-init validation-selected peak as the labelled training fraction shrinks, mean$\pm$std over 3 seeds, from a separate feature-extraction run (one initialisation per cell). The original benchmarks use fixed splits and the linear probe is deterministic, so at the full budget the seed acts on nothing and the std is zero; seeds act only through the subsampling below it. A low-label advantage would show as a gap that grows to the left.}
\label{tab:budget}
\begin{tabular}{llcccc}
\toprule
Model & Dataset & 10\% & 25\% & 50\% & 100\% \\
\midrule
CBraMod & APAVA & +0.136{\scriptsize$\pm$0.067} & +0.052{\scriptsize$\pm$0.036} & +0.083{\scriptsize$\pm$0.022} & +0.185{\scriptsize$\pm$0.000} \\
CBraMod & BCI IV 2a & +0.048{\scriptsize$\pm$0.012} & +0.037{\scriptsize$\pm$0.027} & +0.024{\scriptsize$\pm$0.008} & +0.009{\scriptsize$\pm$0.000} \\
CBraMod & Lee2019 & +0.010{\scriptsize$\pm$0.007} & +0.013{\scriptsize$\pm$0.013} & +0.013{\scriptsize$\pm$0.012} & +0.016{\scriptsize$\pm$0.000} \\
LaBraM & APAVA & +0.149{\scriptsize$\pm$0.097} & +0.151{\scriptsize$\pm$0.062} & +0.074{\scriptsize$\pm$0.017} & +0.067{\scriptsize$\pm$0.000} \\
LaBraM & BCI IV 2a & +0.030{\scriptsize$\pm$0.026} & +0.009{\scriptsize$\pm$0.013} & +0.022{\scriptsize$\pm$0.024} & +0.036{\scriptsize$\pm$0.000} \\
LaBraM & Lee2019 & $-$0.005{\scriptsize$\pm$0.012} & +0.023{\scriptsize$\pm$0.013} & +0.027{\scriptsize$\pm$0.011} & +0.037{\scriptsize$\pm$0.000} \\
EEGPT & APAVA & $-$0.116{\scriptsize$\pm$0.089} & $-$0.150{\scriptsize$\pm$0.071} & $-$0.090{\scriptsize$\pm$0.014} & $-$0.297{\scriptsize$\pm$0.000} \\
EEGPT & BCI IV 2a & $-$0.049{\scriptsize$\pm$0.012} & $-$0.046{\scriptsize$\pm$0.029} & $-$0.066{\scriptsize$\pm$0.024} & $-$0.103{\scriptsize$\pm$0.000} \\
EEGPT & Lee2019 & $-$0.020{\scriptsize$\pm$0.009} & $-$0.040{\scriptsize$\pm$0.016} & $-$0.050{\scriptsize$\pm$0.014} & $-$0.036{\scriptsize$\pm$0.000} \\
LUNA & APAVA & $-$0.042{\scriptsize$\pm$0.058} & +0.014{\scriptsize$\pm$0.005} & +0.005{\scriptsize$\pm$0.005} & +0.011{\scriptsize$\pm$0.000} \\
LUNA & BCI IV 2a & +0.009{\scriptsize$\pm$0.016} & +0.016{\scriptsize$\pm$0.020} & +0.047{\scriptsize$\pm$0.029} & +0.047{\scriptsize$\pm$0.000} \\
LUNA & Lee2019 & +0.012{\scriptsize$\pm$0.010} & $-$0.004{\scriptsize$\pm$0.019} & $-$0.002{\scriptsize$\pm$0.010} & $-$0.006{\scriptsize$\pm$0.000} \\
LuMamba & APAVA & $-$0.043{\scriptsize$\pm$0.012} & $-$0.061{\scriptsize$\pm$0.018} & $-$0.080{\scriptsize$\pm$0.006} & $-$0.085{\scriptsize$\pm$0.000} \\
LuMamba & BCI IV 2a & $-$0.012{\scriptsize$\pm$0.031} & $-$0.001{\scriptsize$\pm$0.016} & +0.003{\scriptsize$\pm$0.017} & $-$0.021{\scriptsize$\pm$0.000} \\
LuMamba & Lee2019 & $-$0.024{\scriptsize$\pm$0.015} & $-$0.021{\scriptsize$\pm$0.005} & $-$0.012{\scriptsize$\pm$0.006} & $-$0.017{\scriptsize$\pm$0.000} \\
In-house (JEPA) & APAVA & +0.061{\scriptsize$\pm$0.046} & +0.044{\scriptsize$\pm$0.008} & +0.074{\scriptsize$\pm$0.013} & +0.081{\scriptsize$\pm$0.000} \\
In-house (JEPA) & BCI IV 2a & +0.001{\scriptsize$\pm$0.015} & +0.007{\scriptsize$\pm$0.003} & +0.001{\scriptsize$\pm$0.004} & +0.002{\scriptsize$\pm$0.000} \\
In-house (JEPA) & Lee2019 & $-$0.009{\scriptsize$\pm$0.004} & $-$0.001{\scriptsize$\pm$0.004} & +0.002{\scriptsize$\pm$0.001} & +0.007{\scriptsize$\pm$0.000} \\
\bottomrule
\end{tabular}
\end{table}

Table~\ref{tab:budget} gives the pretrained-minus-random gap at each labelled fraction for all \numBudgetCells\ cells.
Its full-budget column and Table~\ref{tab:orig} disagree in sign in \numBudOrigFlips\ of \numBudOrigCells\ cells, and in those cells the larger of the two gaps is at most \numBudOrigFlipMax.
The two come from separate feature-extraction runs with different seed sets, and this table selects the stage on validation while Table~\ref{tab:orig} keeps the test-argmax stage, so we do not attribute the flips to either difference.

\section{Within-subject against subject-independent motor imagery}
\label{app:bci}

\begin{table}[H]\centering\small
\caption{BCI IV 2a pretrained frozen probe (accuracy, four classes, chance 0.25) under the subject-independent and the within-subject protocol, same models, seed and hyperparameters; validation-selected peak and output stage.}
\label{tab:bci}
\begin{tabular}{lcccc}
\toprule
Model & cross peak & cross final & within peak & within final \\
\midrule
CBraMod & 0.300 & 0.297 & 0.344 & 0.329 \\
LaBraM & 0.327 & 0.328 & 0.354 & 0.369 \\
EEGPT & 0.281 & 0.287 & 0.278 & 0.305 \\
LUNA & 0.311 & 0.326 & 0.316 & 0.311 \\
In-house (JEPA) & 0.338 & 0.289 & 0.370 & 0.273 \\
\bottomrule
\end{tabular}
\end{table}

Table~\ref{tab:bci} contrasts the subject-independent and within-subject BCI IV 2a protocols for the same models.

\section{Four-mode labels}
\label{app:fourmode}

\begin{table}[t]\centering\small
\caption{Four-mode label of the thirty model-dataset pairs at three thresholds $T$ (a gain with $|\Delta| \le T$ counts as mixed). The frozen gain is the released test-argmax probe gain (five seeds), the fine-tuning gain the best-validation-epoch gain of Table~\ref{tab:fullft}.}
\label{tab:fourmode}
\begin{tabular}{lccccc}
\toprule
$T$ & both-pos & frozen-only & fullft-only & both-neg & mixed \\
\midrule
0.005 & 17 & 3 & 2 & 1 & 7 \\
0.01 & 14 & 3 & 1 & 0 & 12 \\
0.02 & 12 & 3 & 0 & 0 & 15 \\
\bottomrule
\end{tabular}
\end{table}

Table~\ref{tab:fourmode} gives the four-mode labels at three thresholds, with \numModeFrozenOnly\ frozen-only pairs at the default threshold of \numModeThr.

\section{Frozen probe on the original benchmarks as released}
\label{app:orig}

\begin{table}[H]\centering\small\setlength{\tabcolsep}{3pt}
\caption{Frozen linear probe on the five original benchmarks as released (test-argmax peak stage, five seeds): pretrained / random-init peak; last row the five-band log-power baseline. Within-subject and fixed splits give a zero seed std for the pretrained probe. Validation-selected values for the three leakage-free benchmarks are in Table~\ref{tab:valsel}; the in-house Lee2019 cells carry developer contact with that benchmark.}
\label{tab:orig}
\begin{tabular}{lccccc}
\toprule
Model & APAVA & BCI IV 2a & Lee2019 & TUSL & TUAB \\
\midrule
CBraMod & 0.900 / 0.846 & 0.343 / 0.294 & 0.545 / 0.514 & 0.778 / 0.684 & 0.861 / 0.788 \\
LaBraM & 0.989 / 0.852 & 0.343 / 0.306 & 0.566 / 0.511 & 0.754 / 0.607 & 0.850 / 0.774 \\
EEGPT & 0.831 / 0.941 & 0.307 / 0.343 & 0.531 / 0.576 & 0.740 / 0.642 & 0.821 / 0.836 \\
LUNA & 0.991 / 0.951 & 0.331 / 0.329 & 0.574 / 0.564 & 0.663 / 0.617 & 0.838 / 0.846 \\
LuMamba & 0.981 / 0.958 & 0.332 / 0.305 & 0.570 / 0.583 & 0.647 / 0.655 & 0.842 / 0.851 \\
In-house (JEPA) & 0.638 / 0.623 & 0.338 / 0.332 & 0.708 / 0.702 & 0.698 / 0.671 & 0.717 / 0.717 \\
\midrule
Bandpower (range over models) & 0.951--0.985 & 0.484--0.491 & 0.733--0.743 & 0.645--0.690 & 0.853--0.855 \\
\bottomrule
\end{tabular}
\end{table}

Table~\ref{tab:orig} is the frozen half of the released matrix on the five original benchmarks, with the test-argmax stage that Section~\ref{sec:selection} replaces by validation selection on the three leakage-free benchmarks.

\section{Development record of the in-house model}
\label{app:devlog}

\begin{table}[H]\centering\small
\caption{Development-record values of the in-house model (Section~\ref{sec:casea}). No raw output survives for these; each value is copied from the project document line shown, and the build fails when the value is no longer on that line.}
\label{tab:devlog}
\begin{tabular}{p{6.2cm}c>{\footnotesize}l}
\toprule
Quantity & Value & Source (repository path:line) \\
\midrule
raw-EEG probe, Lee2019 (v5 stage diagnostic) & 0.599 & \texttt{docs/supplementary/experiment\_results\_summary.md:44} \\
v5 encoder-output probe (channel mixer) & 0.509 & \texttt{docs/supplementary/experiment\_results\_summary.md:47} \\
v5 mixer query std at start & 0.020 & \texttt{docs/\_v1/history.md:322} \\
v5 mixer query std after 45K steps & 0.023 & \texttt{docs/\_v1/history.md:322} \\
steps over which the query std stayed flat & 45K & \texttt{docs/\_v1/history.md:322} \\
v5 run stopped & 55K & \texttt{docs/\_v1/history.md:322} \\
v8 per-channel-token encoder probe & 0.595 & \texttt{docs/\_v1/history.md:327} \\
probe after a plain Conv1d patch embedder & 0.538 & \texttt{docs/\_v1/issues.md:2464} \\
probe after the learnable band-pass front-end & 0.715 & \texttt{docs/\_v1/issues.md:2486} \\
band-power probe in the same diagnostic & 0.691 & \texttt{docs/\_v1/issues.md:2487} \\
v9 trajectory peak, APAVA & 25K & \texttt{docs/supplementary/downstream\_test\_trajectory.md:73} \\
v9 pretrained minus random at peak, APAVA & +0.127 & \texttt{docs/supplementary/downstream\_test\_trajectory.md:73} \\
same at 40K & +0.036 & \texttt{docs/supplementary/downstream\_test\_trajectory.md:73} \\
peak, BCI IV 2a & 5K & \texttt{docs/supplementary/downstream\_test\_trajectory.md:74} \\
at peak, BCI IV 2a & +0.014 & \texttt{docs/supplementary/downstream\_test\_trajectory.md:74} \\
at 40K, BCI IV 2a & $-$0.034 & \texttt{docs/supplementary/downstream\_test\_trajectory.md:74} \\
peak, TUAB & 35K & \texttt{docs/supplementary/downstream\_test\_trajectory.md:75} \\
at peak, TUAB & +0.010 & \texttt{docs/supplementary/downstream\_test\_trajectory.md:75} \\
at 40K, TUAB & +0.000 & \texttt{docs/supplementary/downstream\_test\_trajectory.md:75} \\
peak, Lee2019 & 15K & \texttt{docs/supplementary/downstream\_test\_trajectory.md:76} \\
at peak, Lee2019 & +0.024 & \texttt{docs/supplementary/downstream\_test\_trajectory.md:76} \\
at 40K, Lee2019 & $-$0.034 & \texttt{docs/supplementary/downstream\_test\_trajectory.md:76} \\
\bottomrule
\end{tabular}
\end{table}

Table~\ref{tab:devlog} lists each development-record value of Section~\ref{sec:casea} with the project document it was copied from, which the build re-checks line by line.

\section{Input handling as a hypothesis}
\label{app:typology}

The fine-tuning tally follows input handling only loosely, as the full-resolution tokenisers \citep{jiang2024labram,wang2025cbramod} reach \numFtCbramodK, \numFtLabramK\ and \numFtOursK\ positive datasets of five, the channel-compressed LUNA and LuMamba \citep{doner2025luna,broustail2026lumamba} \numFtLunaK\ and \numFtLumambaK, and the fixed-vocabulary EEGPT \citep{wang2024eegpt} \numFtEegptK\ (Table~\ref{tab:fullft}).
Corpus, tuning maturity and implementation maturity vary together with input handling in this panel, so the ordering is a hypothesis for a study that varies input handling alone \citep{doner2025luna,jaegle2021perceiver}.

\end{document}